%% file: main.tex
\documentclass[10pt,twocolumn,letterpaper]{article}

\usepackage{cvpr}              

\input{preamble}

\definecolor{cvprblue}{rgb}{0.21,0.49,0.74}
\usepackage[pagebackref,breaklinks,colorlinks,citecolor=cvprblue]{hyperref}
\usepackage{url}
\usepackage{bm}
\usepackage{here}
\usepackage{multirow}

\def\paperID{312} 
\def\confName{3DV\xspace}
\def\confYear{2024\xspace}
\def\methodName{ActiveNeuS\xspace}

\title{\methodName: Neural Signed Distance Fields for Active Stereo}

\author{
Kazuto Ichimaru \quad Takaki Ikeda \quad Diego Thomas \quad Takafumi Iwaguchi \quad Hiroshi Kawasaki \\
Kyushu University, Japan\\
{\tt\small https://www.cvg.ait.kyushu-u.ac.jp/index.html}
}

\begin{document}
\maketitle
\input{sec/0_abstract}    
\input{sec/1_intro}
\input{sec/2_related_work}
\input{sec/3_overview}

\input{sec/4_method}
\input{sec/5_experiments}

\input{sec/6_conclusion}

\vspace{-0.2cm}
\section*{Acknowledgment}
\vspace{-0.3cm}

This work was supported by JSPS KAKENHI Grant Number 
JP20H00611, JP21H01457, JP23H03439.

\clearpage
{
    \small
    \bibliographystyle{ieeenat_fullname}
    \bibliography{main}
}

\end{document}

%% file: preamble.tex
\usepackage[dvipsnames]{xcolor}

\newcommand{\aka}{{\it a.k.a.}}

\newcommand{\fnote}[1]{}

\newcommand{\kcut}[1]{}
\newcommand{\scut}[1]{}
\newcommand{\ocut}[1]{}

%% file: sec/0_abstract.tex
\begin{abstract}

3D-shape reconstruction in extreme environments, such as low illumination or scattering condition, has been an open problem and intensively researched.
Active stereo is one of potential solution for such environments for its robustness and high accuracy.
However, active stereo systems usually consist of specialized system configurations with complicated algorithms, which narrow their application.
In this paper, we propose Neural Signed Distance Field 
for active stereo systems to enable implicit correspondence search and triangulation in generalized Structured Light. 
With our technique, textureless or equivalent surfaces by low light condition are successfully reconstructed even with a small number of captured images.
Experiments were conducted to confirm that the proposed method could achieve state-of-the-art reconstruction quality under such severe condition.
We also demonstrated that the proposed method worked in an underwater scenario. 
\end{abstract}

%% file: sec/1_intro.tex
\section{Introduction}
\label{sec:intro}

For decades, demands for accurate 3D measurement have been growing in the field of autonomous vehicle control, 
industrial inspection, and so on.
Especially, 3D measurement in extreme environments gains more attention,
such as underwater, foggy or minute environments. 
Among those severe environments, textureless or similar conditions under low illumination, low contrast, or heavy scattering effects have been an open problem and active stereo is considered one of the best solution.
However, active stereo often suffers from a correspondence search problem, \ie, finding corresponding points between captured images and projected patterns, and thus,
it usually requires a large number of input images.
Since there is a trade-off between the number of images and the ambiguity of image-pattern correspondence, it becomes more difficult to find accurate correspondences from fewer images, ultimately one image, \aka ~{\it one-shot} scan.
To tackle this problem, many active stereo methods use specialized projection patterns with complicated algorithms, which narrow their applications.

Recently, Neural Radiance Fields (NeRF)~\cite{NeRF} and its variants draw wide attention and brought breakthroughs into many multi-view problems. 
Since they directly optimize Deep Neural Networks (DNN) to minimize a photometric loss in an end-to-end manner, they achieve remarkable accuracy on novel-view synthesis, 3D-shape reconstruction~\cite{NeuS}, super resolution~\cite{wang2021nerf-sr}, etc. 
However, it is known that 3D-shape reconstruction by NeRF or its variants are usually degraded in texture-less/low-texture region under low illumination environment, as their principle is equivalent with conventional passive stereo techniques.

To tackle the low illumination problem, we propose \textbf{\methodName}, which is an active stereo, introducing \textbf{volumetric rendering with pattern projection} into Neural Signed Distance Field (Neural SDF) pipeline.
Our method can be generalized for a wide variation of active stereo systems,
free from correspondence search by implicitly optimizing image-pattern correspondences and scene shape simultaneously, allowing worked even with one-shot scan.
In addition, by \textbf{estimating illumination condition} concurrently 
for each scene, our technique achieved robustness against illumination condition change, which is critical for an actual scanning process.
In the experiments, we show that \methodName achieves accurate one-shot 3D reconstruction in low illumination conditions using synthetic datasets, as well as low-texture scene in real underwater scenario.

Our contributions are as follows:
\begin{itemize}
    \item A novel method for active stereo algorithm using a Neural SDF pipeline is proposed.
    \item Illumination estimation technique is presented to adapt to in-the-wild illumination conditions.
    \item Adequate experiments with both synthetic and real data were conducted to show the feasibility and effectiveness of the proposed method.
\end{itemize}

%% file: sec/2_related_work.tex
\section{Related work}
\label{sec:related_work}

\subsection{Active Stereo}
Active stereo is a methodology to estimate scene shape using active light sources.
In a broad sense, active stereo may include projected pattern stereo~\cite{ProjectedTextureStereo} or photometric stereo~\cite{ActivePhotometricStereo}, but we focus on Structured Light (SL) in this context.
SL uses projected patterns as clues to find correspondences for triangulation.
Most SL methods use dedicated patterns, such as Graycode~\cite{Graycode}, Phase-shifting~\cite{PhaseShift}, Grid-pattern~\cite{Furukawa_2022_WACV}, random dot pattern~\cite{Gu2020sensors}, colorful line pattern~\cite{Fernandez2012icip}, cross laser pattern~\cite{Nagamatsu2021IROS,Nagamatsu2022ICPR}, and so on, to facilitate searching correspondences and reduce number of required images.
Some SL methods optimize patterns by themselves to minimize ambiguity of correspondences and maximize useful information~\cite{Parsa2018cvpr,chen_2020_autotuning}.
There are also commercial depth cameras using SL methods~\cite{Kinect,RealSense}.

Recently, Differentiable Rendering (DR) draws wide attention for its effectiveness of end-to-end optimization.
Some use DR to optimize pattern~\cite{baek2021polka}, while others use DR as inverse problem solvers to directly estimate scene as depth image~\cite{Benjamin2021DDS}, or triangle mesh~\cite{Janus2021SLDR}.
DR is also closely related to NeRF, which is mentioned in the next section.

\subsection{Neural Radiance Fields}
Neural Radiance Fields (NeRF) is a methodology to represent a scene as a volumetric function which outputs density of a 3D point and color from a specific viewing direction~\cite{NeRF}.
NeRF utilizes DNN's strong power of interpolation and extrapolation to accurately generate novel views from limited images, or estimate scene shape.
Some use NeRF as a BRDF tool for albedo acquisition or re-lighting~\cite{NeRD}.
While original NeRF performs well on novel view synthesis, its performance on 3D-shape reconstruction is limited since volumetric density function is not suitable to represent solid surfaces.
VolSDF and NeuS replaced volumetric density function with Signed Distance Function (SDF), which outputs signed distance to the closest surface~\cite{VolSDF,NeuS}.
Neural SDF drastically improved 3D-shape reconstruction accuracy by introducing inductive bias.

One of the major limitations of NeRF and its variants is the number of required images to learn a scene.
Depth supervision is known to be effective to reduce the number of required images.
DS-NeRF uses depth obtained from conventional Structure-from-Motion (SfM) as extra supervision to successfully learn a scene from a small (2-10) number of images~\cite{DSNeRF}.
However, accurate depth data is usually expensive and depth from SfM often contains large errors in texture-less region.
A combination of NeRF and active stereo will be a remedy for such a problem.

\subsection{NeRF variants for Active Stereo}
Although active stereo and NeRF have compatibility in terms of DR, there are few attempts to combine NeRF with active stereo.
Li \etal combined Graycode SL~\cite{Graycode} with Neural SDF (NeuS~\cite{NeuS}) to cope with an inter-reflection problem, which severely degrades reconstruction quality~\cite{Chunyu2022DRSL}.
However, their data acquisition process must be precisely controlled in the laboratory environment since Graycode SL requires multiple projections from a single viewpoint.
Furthermore, they explicitly compute image-pattern correspondences via a naive Graycode SL algorithm, but it becomes unstable in extreme environments.
Since the proposed method implicitly computes image-pattern correspondences via volumetric rendering with pattern projection, it is generalized to multi-view multi-shot SL with arbitrary projection patterns.
This will be the first attempt to achieve static pattern based SL with Neural SDF with the best of our knowledge. 
Since the pattern is static, it is possible to capture the scene with dynamically moving the system as well.
\autoref{tab:diff} shows differences between the proposed method and previous methods.

\begin{table}[t]
    \centering
    \caption{Differences between the proposed method and previous methods.}
    \label{tab:diff}
    \vspace{-0.3cm}
    \begin{tabular}{p{2.2 cm}|p{1.4 cm}|p{1.6 cm}|p{1.6 cm}}
        \hline
        Methods & NeuS~\cite{NeuS} & DR\&SL~\cite{Chunyu2022DRSL} & \methodName \\ \hline\hline
        Reconstruction algorithm & Passive Stereo & Graycode & \textbf{Generalized SL} \\ \hline
        Correspondence search & - & Explicit & \textbf{Implicit} \\ \hline
        \# of required images & Medium ($\ge20$) & Medium ($\ge20$) & \textbf{Small ($\ge1$)} \\ \hline
    \end{tabular}
\end{table}

%% file: sec/3_overview.tex
\section{Method Overview}
\label{sec:method_overview}


\begin{figure}[t!]
    \centering
    \includegraphics[width=0.8\linewidth]{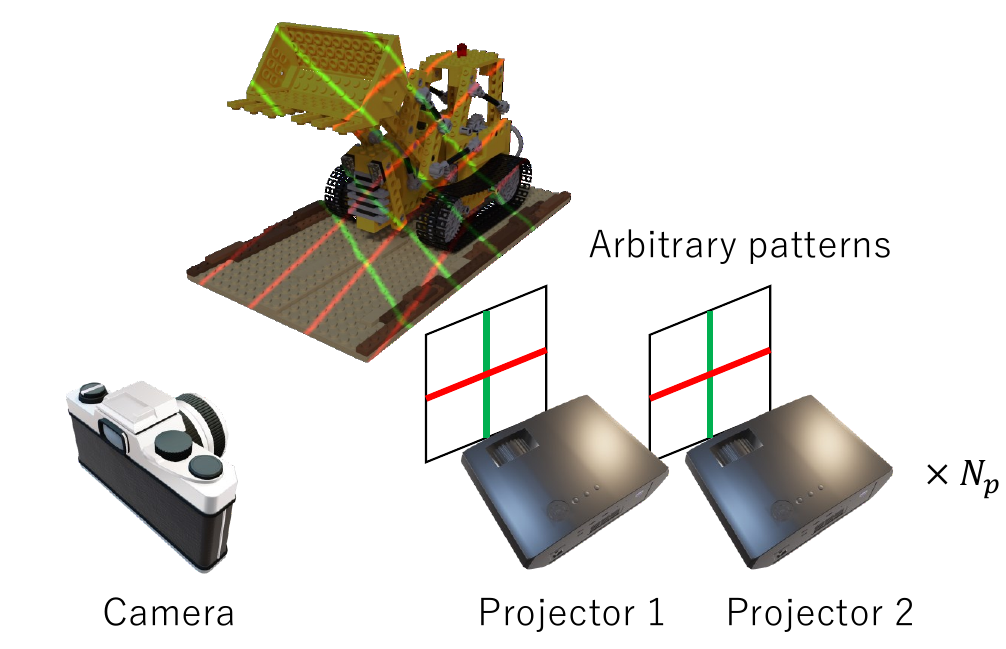}
    \vspace{-0.3cm}
    \caption{System configuration.}
    \label{fig:system_configuration}
\end{figure}

\begin{figure*}[t!]
    \centering
    \includegraphics[width=\linewidth]{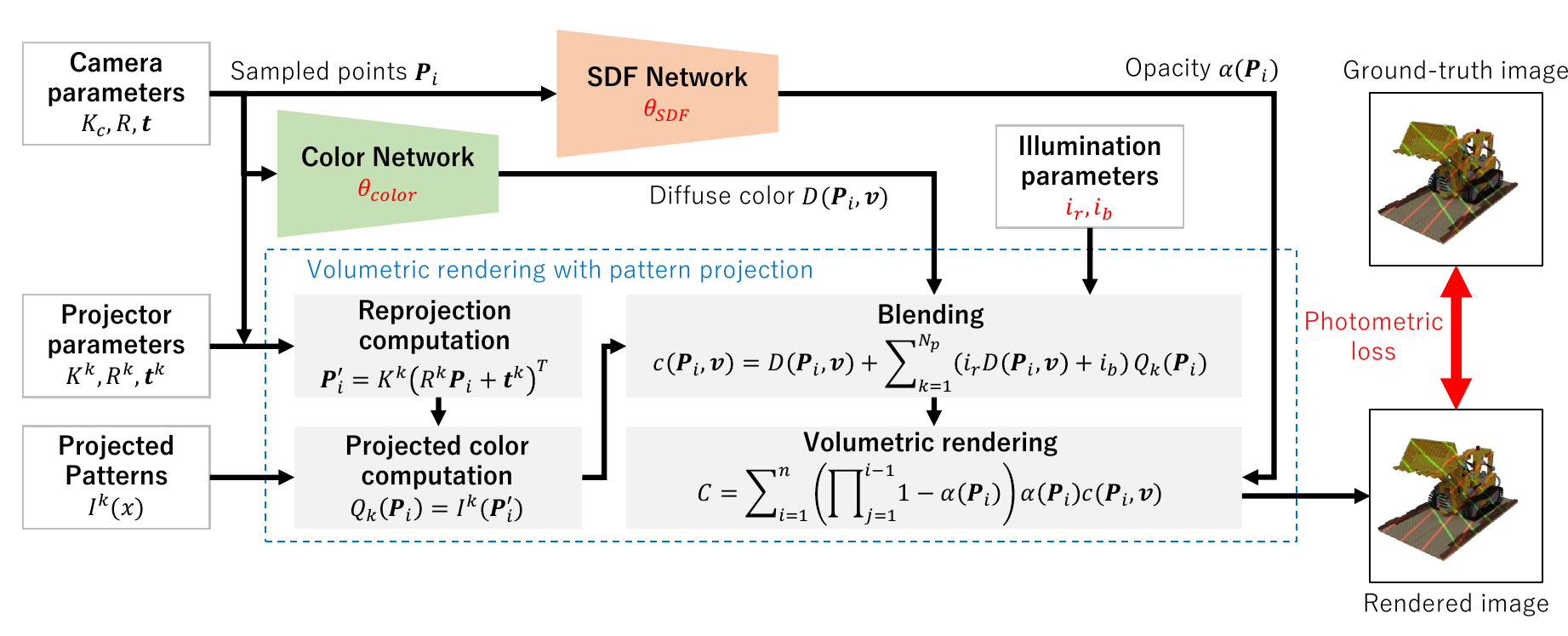}\\
    \vspace{-0.5cm}
    \caption{Pipeline schematics of the proposed method. Parameters marked in red are optimized during training.}
    \label{fig:pipeline}
\end{figure*}

\subsection{System configuration and environmental assumption}

The proposed method assumes system configuration with a camera and an arbitrary number of projectors whose intrinsic and extrinsic parameters are accurately calibrated, as shown in \autoref{fig:system_configuration} ($N_p$ denotes the number of projectors).
Relative transformations between the camera and the projectors can vary during capturing as long as their extrinsic parameters can be retrieved.
The projectors project arbitrary patterns, such as Graycode, Grid-pattern, Cross-laser pattern, and so on.
The patterns do neither have to be static during capturing, nor consistent across the projectors in theory, but we used static and consistent patterns (mainly Cross-laser pattern) in the experiments.
As for the environment, we assume the scenes are static during capturing, and Lambertian reflectance is dominant.

\subsection{Pipeline of the proposed method}

The proposed method mainly follows NeuS\cite{NeuS} pipeline, which consists of the SDF network and the Color network, except that our pipeline has \textbf{Projector parameters}, \textbf{Projected patterns} and  \textbf{Illumination parameters} which consist \textbf{Volumetric rendering with pattern projection} branch.
\autoref{fig:pipeline} shows the pipeline of the proposed method.
Note that we omit some modules from the figure used in the pipeline, such as hierarchical sampling and variance network, which are also used by NeuS.

The training procedure of the pipeline is as follows.
\begin{enumerate}
    \item Randomly sample rays casting from each camera's optical centers using camera parameters.
    \item Sample 3D points on the ray from the near clip to the far clip at regular or weighted intervals.
    \item Pass the 3D points to the SDF / Color networks to acquire the density and diffuse color of the points.
    \item Re-project the 3D points onto the projector patterns to get 2D projected points using projector parameters.
    \item Compute projected colors to the 3D points from the projected patterns using the 2D projected points.
    \item Render images by \textbf{volumetric rendering with pattern projection}.
    \item Compute photometric loss between the rendered images and the Ground-truth images.
    \item Update the network parameters to minimize the photometric loss using Adam.
\end{enumerate}

By updating the network parameters, SDF is optimized to minimize the discrepancy between the projected pattern and the Ground-truth image, \ie implicitly searching image-pattern correspondences.
Textures are also implicitly utilized to learn more detailed SDF.

%% file: sec/4_method.tex
\section{Neural SDF for Active Stereo}
\label{sec:method}


\subsection{Volumetric rendering with pattern projection}

In ordinary Neural SDF, a 2D point $p$ is rendered using 3D points $\bm{P}_i$ ($i=1..n$) on the ray cast from $p$ as \autoref{eq:volumetric_rendering}
\begin{equation}
    C = \sum_{i=1}^n \left( \prod_{j=1}^{i-1}1-\alpha(\bm{P}_j) \right) \alpha(\bm{P}_i) c(\bm{P}_i, \bm{v}),
    \label{eq:volumetric_rendering}
\end{equation}
where $\alpha(\bm{P}_i)$ is opacity of a 3D point $\bm{P}_i$, $c(\bm{P}_i, \bm{v})$ is color of $\bm{P}_i$ viewing from direction $\bm{v}$, and $C$ is final rendered color of the ray.
Following NeuS, $\alpha(\bm{P}_i)$ is computed as \autoref{eq:alpha},
\begin{equation}
    \alpha(\bm{P}_i) = \max\left(\frac{\Phi_s(f(\bm{P}_i)) - \Phi_s(f(\bm{P}_{i+1}))}{\Phi_s(f(\bm{P}_i))}, 0\right)
    \label{eq:alpha}
\end{equation}
where $f(\bm{P})$ is SDF value of a 3D point $\bm{P}$ and $\Phi_s$ is Sigmoid function in our implementation.

Usually, $c(\bm{P}_i,\bm{v})$ is the Color network itself, but we modify it to compute color with pattern projection.
We blend diffuse color of $\bm{P}_i$ (denoted as $D(\bm{P}_i,\bm{v})$) and projected color by $k$-th projector (denoted as ${Q_k(\bm{P}_i)}$) as \autoref{eq:color_blend},
\begin{equation}
    c(\bm{P}_i,\bm{v}) = D(\bm{P}_i,\bm{v}) + \sum_{k=1}^{N_p} (i_{r}D(\bm{P}_i,\bm{v}) + i_{b})Q_k(\bm{P}_i),
    \label{eq:color_blend}
\end{equation}
where $i_r$ is reflectance coefficient, and $i_b$ is bias coefficient.
$i_r$ controls reflectance level of $\bm{P}_i$, and $i_b$ controls emissive level of $Q(\bm{P}_i)$.
By properly setting $i_r$ and $i_b$, we can render a high-fidelity image with pattern projection.

The intuition of this equation is that, if we can render an image with pattern projection correctly, it means we have an accurate depth and texture, and vice versa.
Thus, it implicitly works as depth supervision, which helps better 3D shape optimization, especially in texture-less regions or low illumination environments which lack information for depth supervision.

In our implementation, $D(\bm{P}_i,\bm{v})$ is the Color network itself, and $Q_k(\bm{P}_i)$ is computed as \autoref{eq:projector_projection},
\begin{equation}
    Q_k(\bm{P}_i) = I^k(K^k(R^k\bm{P}_i + \bm{t}^k)^T)
    \label{eq:projector_projection}
\end{equation}
where $I^k(x)$ returns the color of the pattern of $k$-th projector at specific point $x$ via bilinear sampling, $K^k$ is the intrinsic matrix and $R^k, \bm{t}^k$ are the relative transformation of $k$-th projector, which convert 3D points in the world coordinate system into the projector screen coordinate system.

\subsection{Illumination estimation}

One remaining problem is how to set illumination parameters $i_r$ and $i_b$.
We empirically observed that slight differences in the parameters from Ground-truth have little impact on reconstruction results, but illumination conditions may change drastically in-the-wild scenes.

As a solution, we include $i_r$ and $i_b$ in optimized parameters during training.
However, optimizing $i_r$ and $i_b$ from the beginning of the training sometimes breaks the SDF network, because it is impossible to estimate illumination conditions without properly rendered images.
We noticed that the rendered images look fine even when the 3D shape is rough.
Thus, we start optimizing $i_r$ and $i_b$ after training for a while to warm up the SDF network.
Incidentally, we insert ReLU layers after $i_r$ and $i_b$ parameters to prevent them from being negative values.

\subsection{Texture retrieval}
Texture retrieval is another major problem of SL systems.
Since people usually want scene texture without pattern projection, they capture the scene without projection separately or remove patterns from the captured images in One-shot case.
In our pipeline, the projected color is blended using illumination parameters $i_r$ and $i_b$.
We can override these parameters to $0$ after training the networks, to retrieve scene texture without projection in the same way as the ordinary novel-view synthesis process of NeRF.

\subsection{Implementation details}

We implemented the proposed method, namely \methodName, following NeuS implementation.
We used hierarchical sampling~\cite{NeRF}, Eikonal loss~\cite{Eikonal}, and mask loss~\cite{NeuS} as well for better efficiency and accuracy.
Eikonal loss is a well known regularizer, which helps learning a spatially consistent SDF. 
As for the occlusion of pattern projection, we found that we can just ignore it and let the SDF network learn forward-backward relation thanks to NeuS's occlusion-aware weighting function design.
Overall, the objective function of the pipeline is as \autoref{eq:objective}
\begin{equation}
    \mathcal{L} = \mathcal{L}_{color} + \lambda\mathcal{L}_{reg} + \beta\mathcal{L}_{mask},
    \label{eq:objective}
\end{equation}
where $\mathcal{L}_{color}$ is photometric loss (L1), $\mathcal{L}_{reg}$ is Eikonal term, $\mathcal{L}_{mask}$ is mask loss term, and $\lambda, \beta$ are balancing coefficients.
Although we implemented mask loss following NeuS implementation, we did not use any mask supervision in the experiments for evaluations in a challenging condition, except in \autoref{ssec:various_patterns}.

As for the hyper-parameters of training, we used learning rate $5e-4$, learning rate decay coefficient $0.05$, batch size $512$, and Eikonal term coefficient $\lambda = 0.1$.
We trained the networks for 200k epochs in the experiments.

%% file: sec/5_experiments.tex
\section{Experiments}
\label{sec:experiments}


\subsection{Quantitative and qualitative evaluations with synthetic data}

We conducted several experiments with synthetic data in some respects.
Throughout the experiments, we used two datasets.

\begin{itemize}
    \item NeRF-Synthetic\cite{NeRF}: Sequences with a single object captured by a camera moving along an orbital trajectory. We used "Lego", "Chair", "Hotdog", and "Mic" sequences.
    \item BlendedMVS\cite{yao2020blendedmvs}: Sequences of various scenes including aerial scenery, sculptures, potteries, and so on. We used the same scenes as NeuS ("Stone", "Dog", "Bear", "Sculpture").
\end{itemize}

\begin{figure}[t!]
    \centering
    \hspace{-0.3cm}
    \begin{minipage}[b]{0.21\linewidth}
        \centering
        \includegraphics[width=\columnwidth]{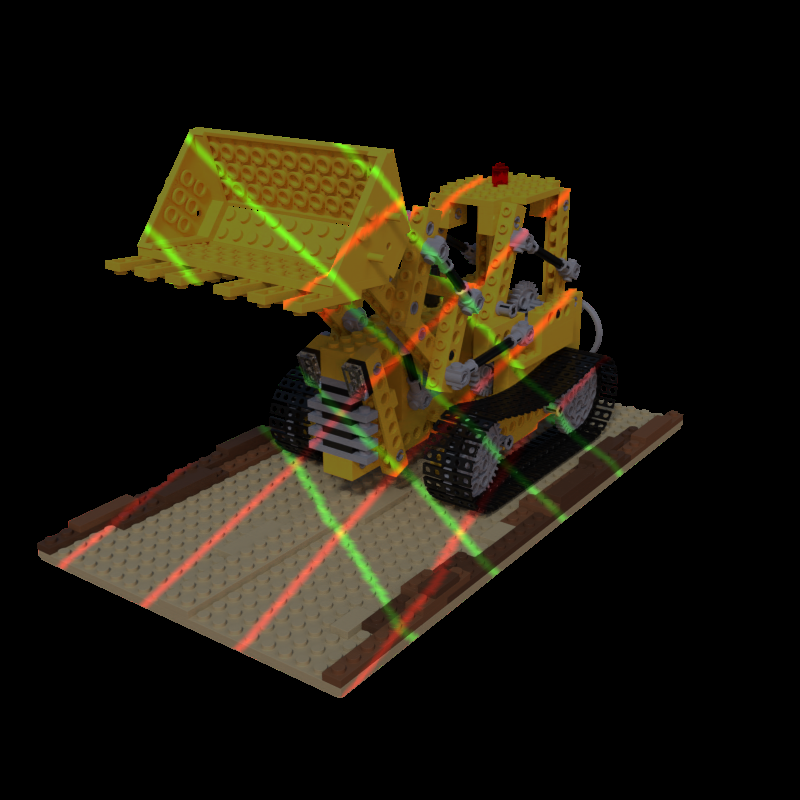}
    \end{minipage}
    \begin{minipage}[b]{0.21\linewidth}
        \centering
        \includegraphics[width=\columnwidth]{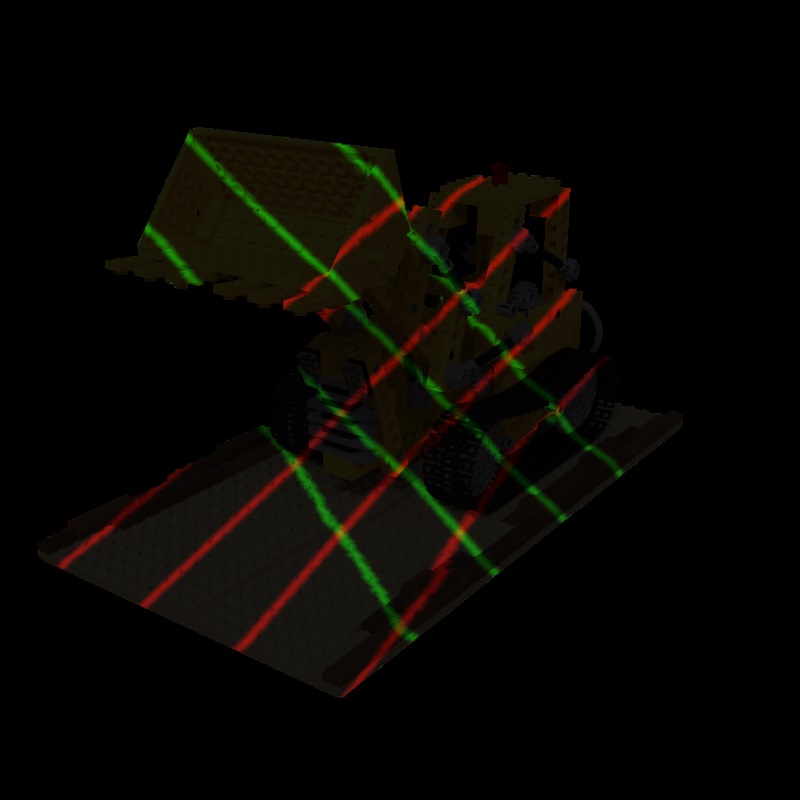}
    \end{minipage}
    \begin{minipage}[b]{0.28\linewidth}
        \centering
        \includegraphics[width=\columnwidth]{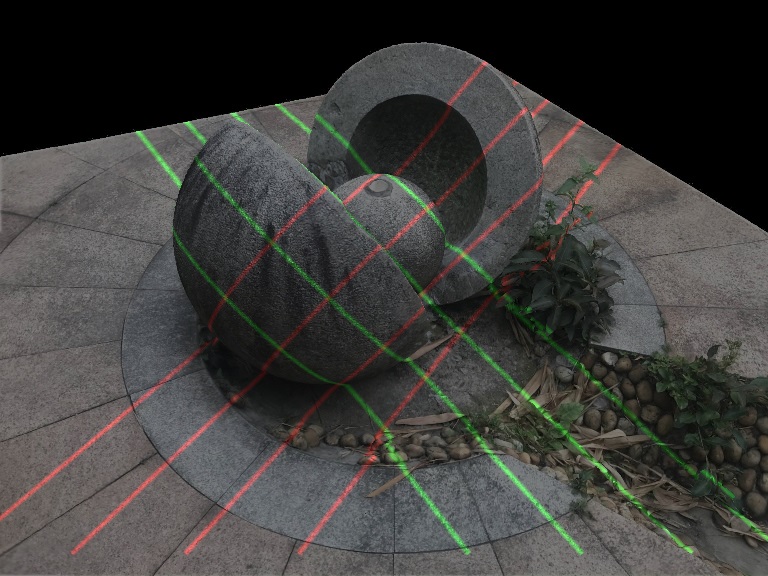}
    \end{minipage}
    \begin{minipage}[b]{0.28\linewidth}
        \centering
        \includegraphics[width=\columnwidth]{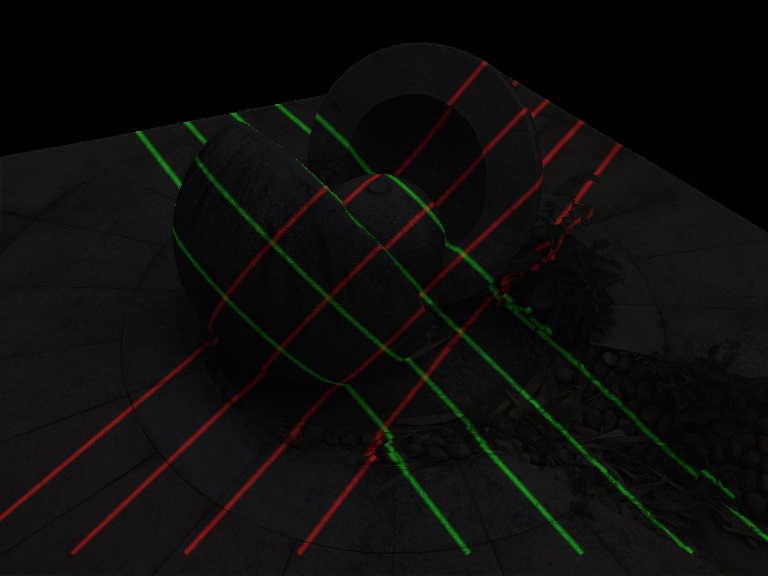}
    \end{minipage}
    \vspace{-0.2cm}
    \caption{Example images of the synthetic data with pattern projection. \textbf{Left}: NeRF-Synthetic (Normal and Dark). \textbf{Right}: BlendedMVS (Normal and Dark). Note that only laser curves are visible for human eyes for Dark illumination scenario.}
    \label{fig:dataset_image}
\end{figure}

For both datasets, we synthesized images with pattern projection by computing rays from virtual projectors.
We used 4 virtual projectors lined up on the left and right of the camera with static Cross-laser patterns (red and green) unless specifically mentioned.
The projectors are assumed to be fixed with the camera at a certain relative rotation and position, \ie the projectors move with the camera.
Baseline lengths of the projectors are $20cm$ and $60cm$ from the camera, and field-of-view is $60^{\circ}$.
\autoref{fig:dataset_image} shows example images of the synthetic data with pattern projection.

\subsubsection{Evaluation on reconstruction accuracy}

\begin{figure*}[t!]
    \centering
    \includegraphics[width=0.9\linewidth]{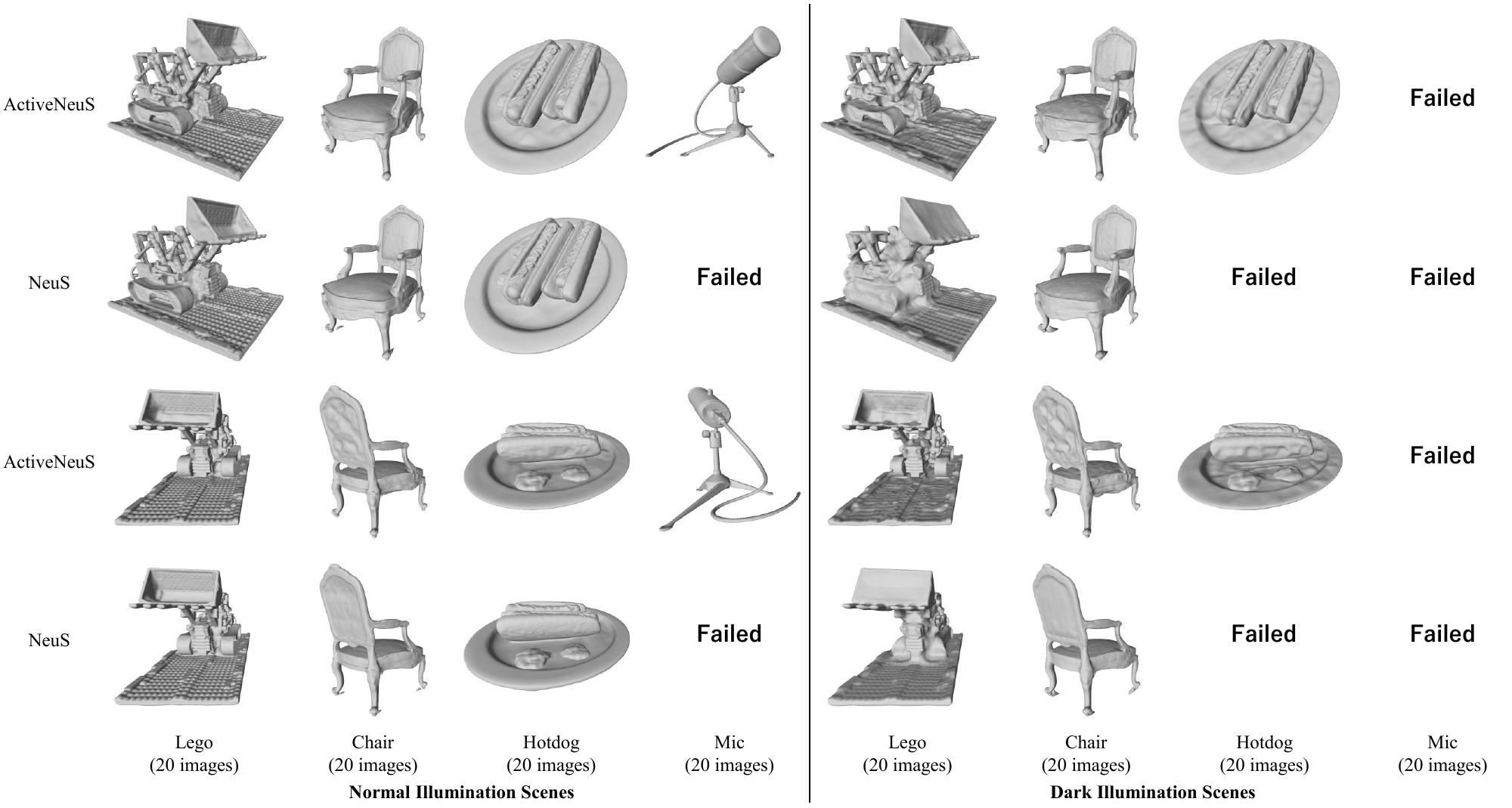}
  \vspace{-0.2cm}
    \caption{Comparison of the reconstructed shapes on NeRF-Synthetic dataset. 20 images are used for input.}
    \label{fig:accuracy_nerf_synthetic}
\end{figure*}

\begin{figure*}[t!]
    \centering
    \includegraphics[width=0.9\linewidth]{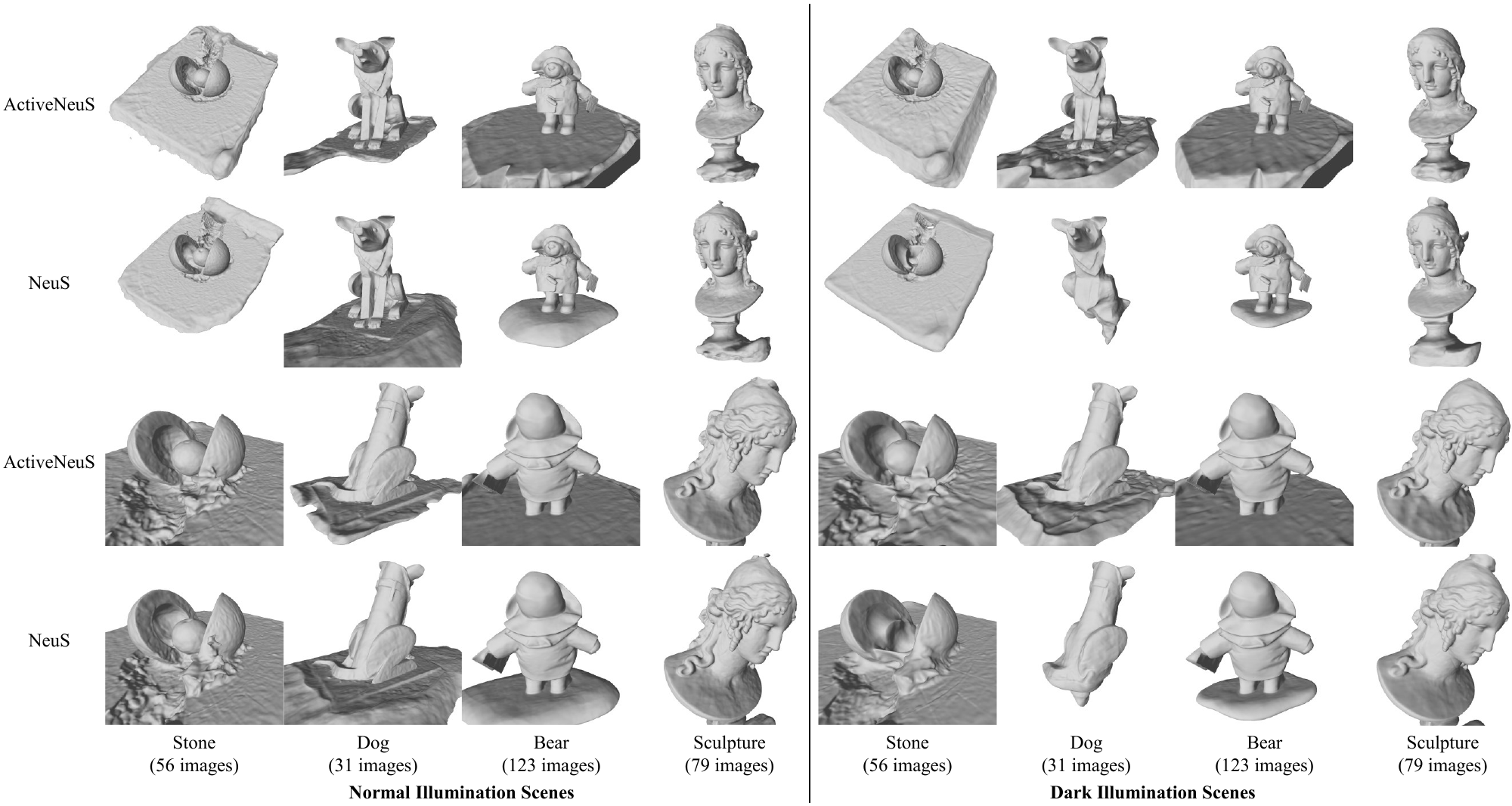}
  \vspace{-0.2cm}
    \caption{Comparison of the reconstructed shapes on BlendedMVS dataset. All images are used for input.}
    \label{fig:accuracy_blendedmvs}
\end{figure*}

In order to evaluate the reconstruction accuracy of the proposed method, we compared reconstructed shapes with NeuS using NeRF-Synthetic and BlendedMVS datasets (DR\&SL~\cite{Chunyu2022DRSL} is specialized for Graycode reconstruction and their main focus is inter-reflection, thus, it is complementary to ours. Similarly, we omit DS-NeRF~\cite{DSNeRF} since it is dedicated to novel-view-synthesis).
As for the NeRF-Synthetic dataset, we uniformly sampled 20 images for each sequence and synthesized images with pattern projection in two illumination conditions by ambient level (Normal and Dark).
As for the BlendedMVS dataset, we used all images for each scene (31-123 images) and synthesized images with pattern projection as same as NeRF-Synthetic.
We omitted pattern projection for NeuS as it may be a severe disturbance for the passive stereo based method.

\autoref{fig:accuracy_nerf_synthetic} and \autoref{fig:accuracy_blendedmvs} show the comparison of the reconstructed shapes (``Failed'' indicates we could not get meaningful results).
As we can see, accurate reconstruction is possible with both NeuS and the proposed method in the Normal illumination condition, while the proposed method is advantageous in the Dark illumination condition.
Some reconstructed shapes are over-smoothed or complete failures with NeuS in the Dark illumination condition, while ours successfully compensated such errors.
As for the Mic scene from NeRF-Synthetic in the Dark illumination condition, both the proposed method and NeuS failed to reconstruct the shape, because the scene contains thin areas with little black texture, and the pattern projection did not provide enough depth supervision.
Unfortunately, \methodName produced bumpier results for some scenes.
We consider it is due to sparse depth supervision by pattern projection.

\autoref{tab:quantitative_nerf_synthetic} and \autoref{tab:quantitative_blendedmvs} show quantitative comparison of the reconstructed shapes.
The metric is chamfer distance from the Ground-truth shapes to the reconstructed shapes in millimeters.
We removed some points from the computation whose distances are larger than $1\sigma$ of all points to ignore ambiguous regions such as the table in BlendedMVS (Bear).
The proposed method achieves comparative or even higher accuracy than NeuS, especially in the Dark illumination condition.

\begin{table}[]
    \centering
    \small
    \caption{Results of quantitative evaluation on NeRF-Synthetic in Chamfer distance [mm]. Entries with ``-'' indicate the cases failed meaningful reconstruction.}
    \label{tab:quantitative_nerf_synthetic}
    \vspace{-0.3cm}
    \begin{tabular}{|cc|cccc|}
        \hline
        Illum & Method & Lego & Chair & Hotdog & Mic \\ \hline \hline

        \multirow{2}{*}{\hspace{-1mm}Normal} & NeuS & 9.1 & \textbf{3.4} & \textbf{5.6} & - \\
        & \hspace{-2mm}\methodName (ours) & \textbf{8.3} & 3.6 & 6.4 & \textbf{1.2} \\ \hline
        \multirow{2}{*}{\hspace{-2mm}Dark} & NeuS & 15.4 & 4.1 & - & - \\
        & \hspace{-2mm}\methodName (ours) & \textbf{7.5} & 4.1 & \textbf{10.0} & - \\ \hline

    \end{tabular}
\end{table}

\begin{table}[]
    \centering
    \small
    \caption{Results of quantitative evaluation on BlendedMVS in Chamfer distance [mm].}
    \label{tab:quantitative_blendedmvs}
    \vspace{-0.3cm}
    \begin{tabular}{|cc|cccc|}
        \hline
        Illum & Method & Stone & Dog & Bear & Sculpture \\ \hline \hline
 
        \multirow{2}{*}{Normal} & NeuS & 6.2 & \textbf{1.0} & 12.9 & 1.3 \\
        & \methodName & \textbf{5.0} & 1.4 & \textbf{2.5} & \textbf{1.2} \\ \hline
        \multirow{2}{*}{Dark} & NeuS & \textbf{6.9} & 8.7 & 21.0 & 1.7 \\
        & \methodName & 7.8 & \textbf{1.3} & \textbf{2.7} & \textbf{1.5} \\ \hline

    \end{tabular}
\end{table}

\subsubsection{Evaluation on illumination estimation}

\begin{figure}[t!]
    \centering
    \begin{minipage}[b]{0.42\linewidth}
        \centering
        \includegraphics[width=\columnwidth]{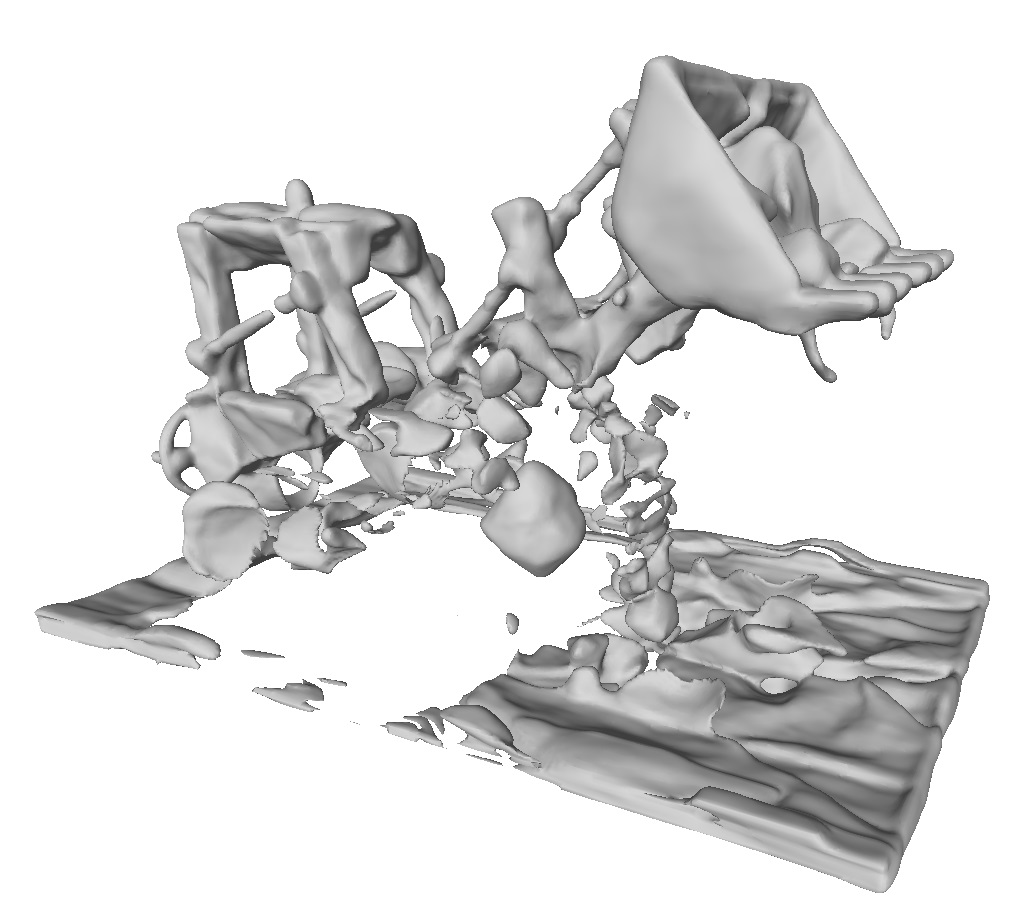}
        \subcaption{w/o Illumination estimation}
    \end{minipage}
    \begin{minipage}[b]{0.42\linewidth}
        \centering
        \includegraphics[width=\columnwidth]{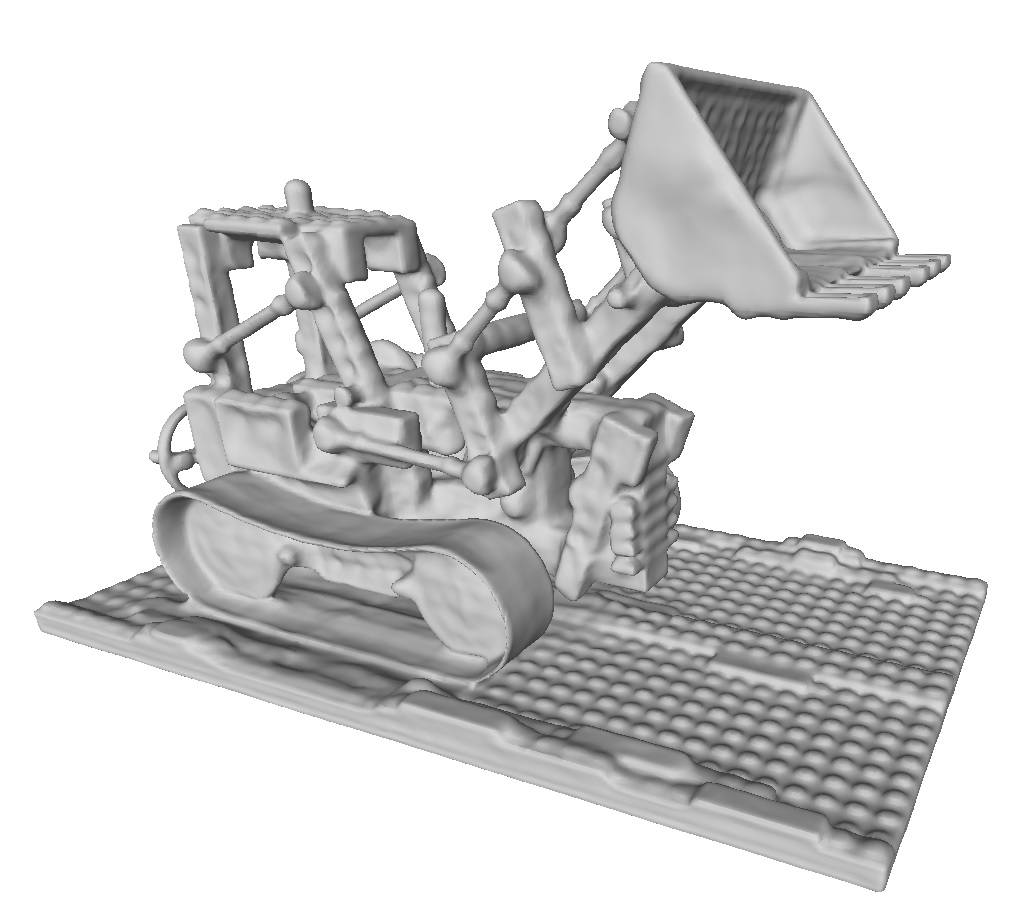}
        \subcaption{w Illumination estimation}
    \end{minipage}
    \caption{Comparison of the reconstructed shapes with and without illumination estimation.}
    \label{fig:illumination_estimation}
\end{figure}

In order to confirm the effectiveness of illumination estimation, we compared reconstructed shapes with and without illumination estimation.
We intentionally changed initial illumination parameters as shown in \autoref{tab:illumination_estimation} and trained the networks with NeRF-Synthetic scenes (Lego, Chair, Hotdog and Mic).
\autoref{fig:illumination_estimation} shows the comparison of the reconstructed shapes (Lego).
As we can see, reconstruction is successful with illumination estimation, while large errors exist without illumination estimation.
The estimated illumination parameters are on \autoref{tab:illumination_estimation}, showing accurate illumination parameters were estimated, which also leads to accurate reconstruction.

\begin{table}[]
    \centering
    \small
    \caption{Results of evaluation on illumination estimation (average on 4 scenes). Shape error is Chamfer distance [mm].}
    \label{tab:illumination_estimation}
    \vspace{-0.3cm}
    \begin{tabular}{|c|c|c|c|c|}
        \hline
         & $i_r$ & $i_b$ & Shape \\ \hline \hline
        Initial & 1.0 & 1.0 & 17.2 ($\pm 14.3$) \\ \hline
        \multirow{2}{*}{After estimation} & 0.403 & 0.0893 & \textbf{4.4} \\
        & ($\pm 0.023$) & ($\pm 0.004$) & ($\pm 0.6$) \\ \hline
        Ground-truth & 0.4 & 0.1 & 4.9 ($\pm 2.7$) \\ \hline
    \end{tabular}
\end{table}

\vspace{-0.2cm}
\subsubsection{Evaluation on texture retrieval}

\begin{figure}[t!]
    \centering
    \includegraphics[width=\columnwidth]{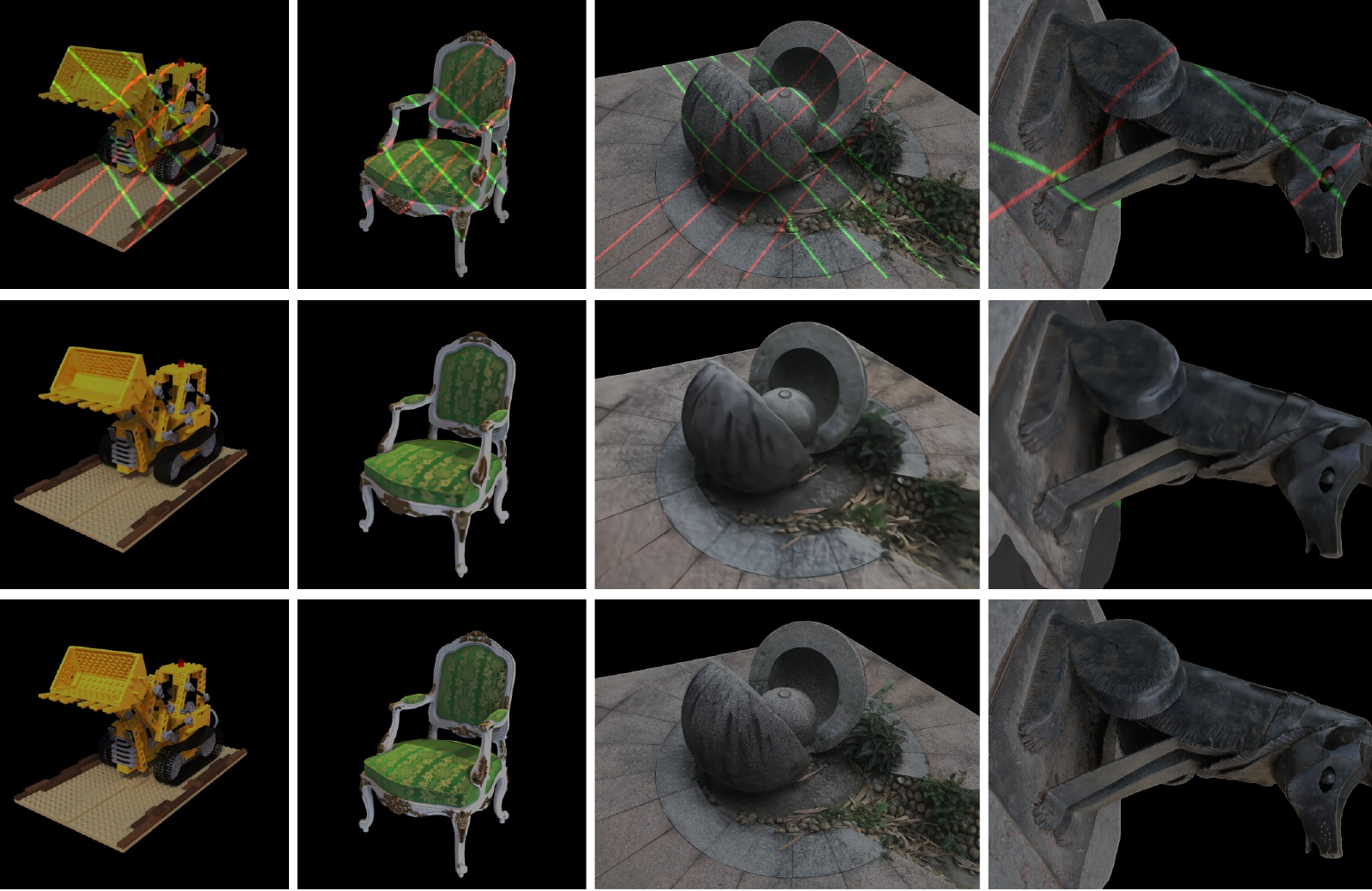}
    \caption{Texture retrieval results. \textbf{Top}: Training images. \textbf{Middle}: Texture removal results. \textbf{Bottom}: Ground-truth images. \textbf{Left to right}: NeRF-Synthetic (Lego, Chair), BlendedMVS (Stone, Dog).}
    \label{fig:texture_retrieval}
\end{figure}

As for texture retrieval, we conducted experiments on NeRF-Synthetic (Lego, Chair) and BlendedMVS (Stone, Dog).
\autoref{fig:texture_retrieval} shows the results of texture retrieval.
Qualitatively, it is shown that accurate texture retrieval is possible with the proposed method with few artifacts.
Unfortunately, some areas were smoothed too much due to texture retrieval, which is a minor limitation of the proposed method.
Note that we aligned the viewpoints for comparison, but texture retrieval is possible for arbitrary viewpoints as NeRF enables novel-view synthesis.

\vspace{-0.2cm}
\subsubsection{Evaluation on pattern variation}
\label{ssec:various_patterns}

\begin{figure}[t!]
    \centering
    \begin{minipage}[b]{0.28\linewidth}
        \centering
        \includegraphics[width=\columnwidth]{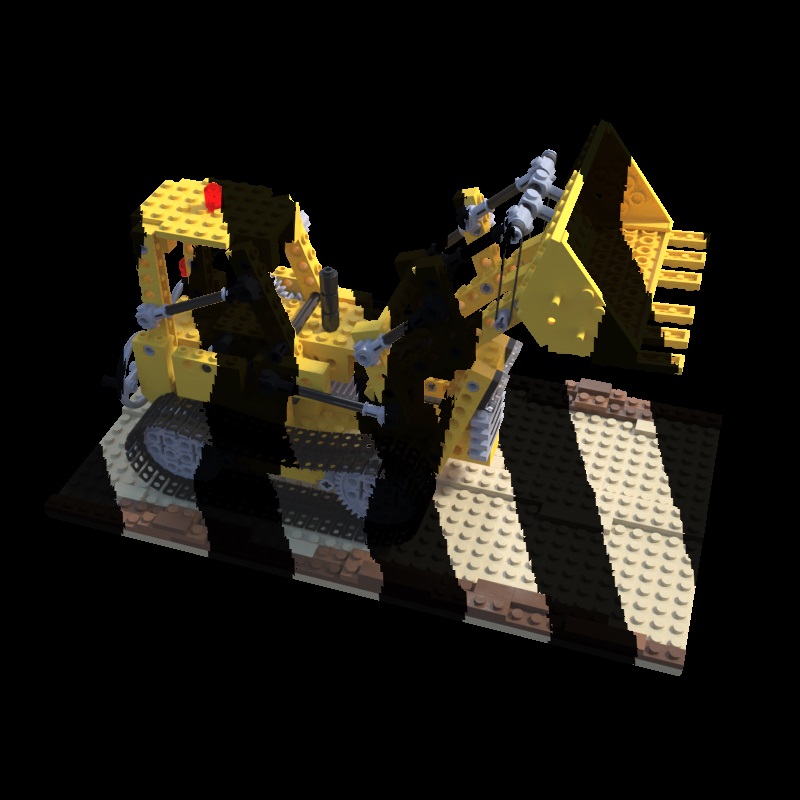}
        \subcaption{Graycode}
    \end{minipage}
    \begin{minipage}[b]{0.28\linewidth}
        \centering
        \includegraphics[width=\columnwidth]{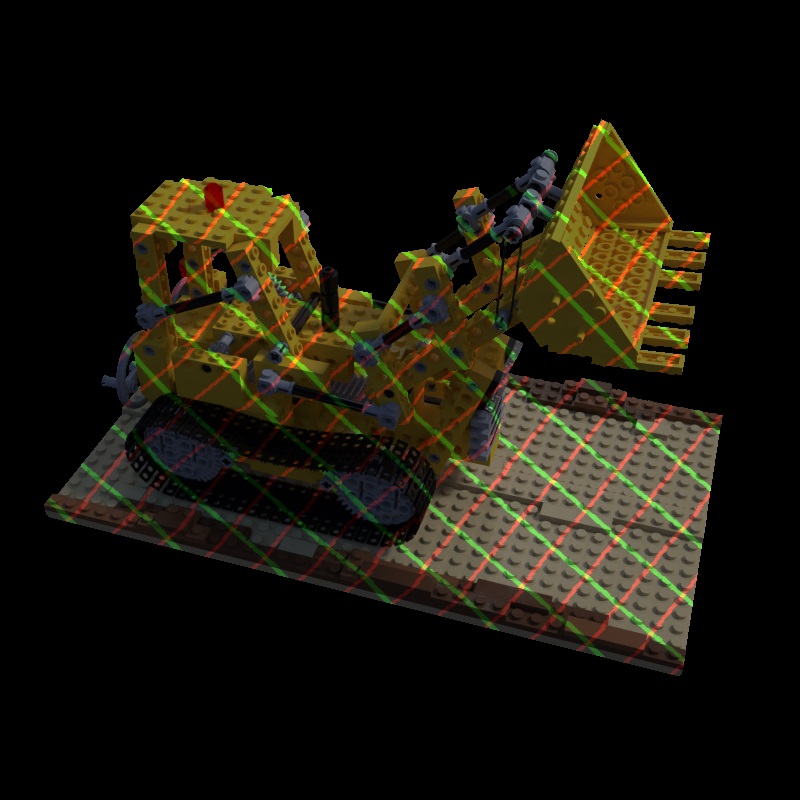}
        \subcaption{Grid-pattern}
    \end{minipage}
    \begin{minipage}[b]{0.28\linewidth}
        \centering
        \includegraphics[width=\columnwidth]{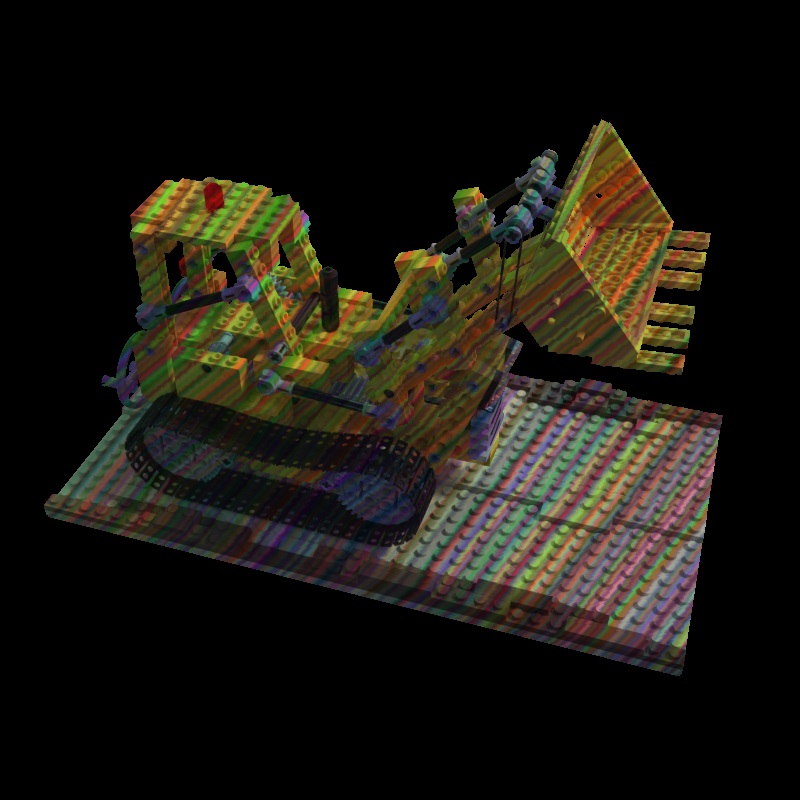}
        \subcaption{Colorful line}
    \end{minipage}
    \caption{Example images of the synthetic data with various patterns.}
    \label{fig:dataset_various_patterns}
\end{figure}

\begin{figure}[t!]
    \centering
    \begin{minipage}[b]{1\linewidth}
        \centering
        \includegraphics[width=\columnwidth]{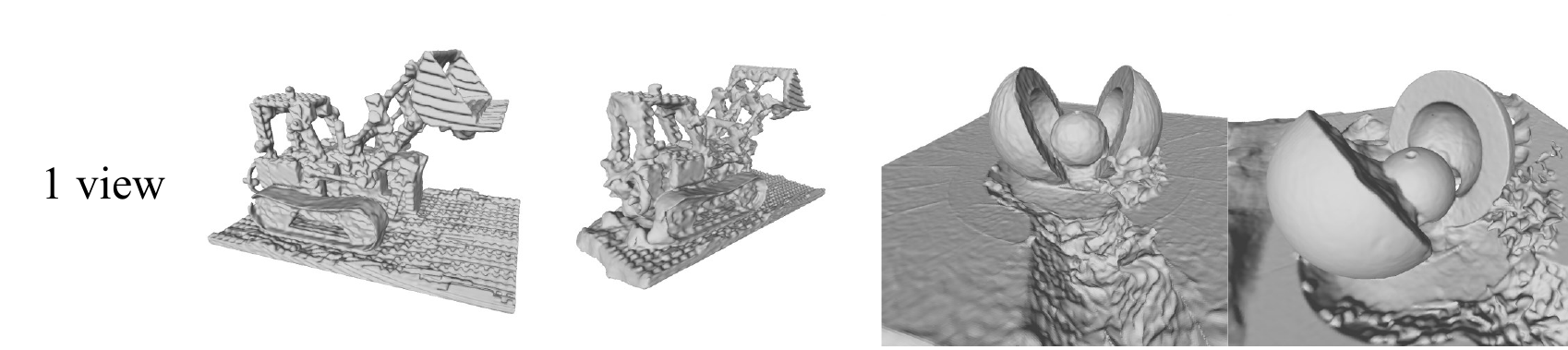}
        \subcaption{Graycode (20 patterns per view)}
 \vspace{0.1cm}
    \end{minipage}
    \begin{minipage}[b]{1\linewidth}
        \centering
        \includegraphics[width=\columnwidth]{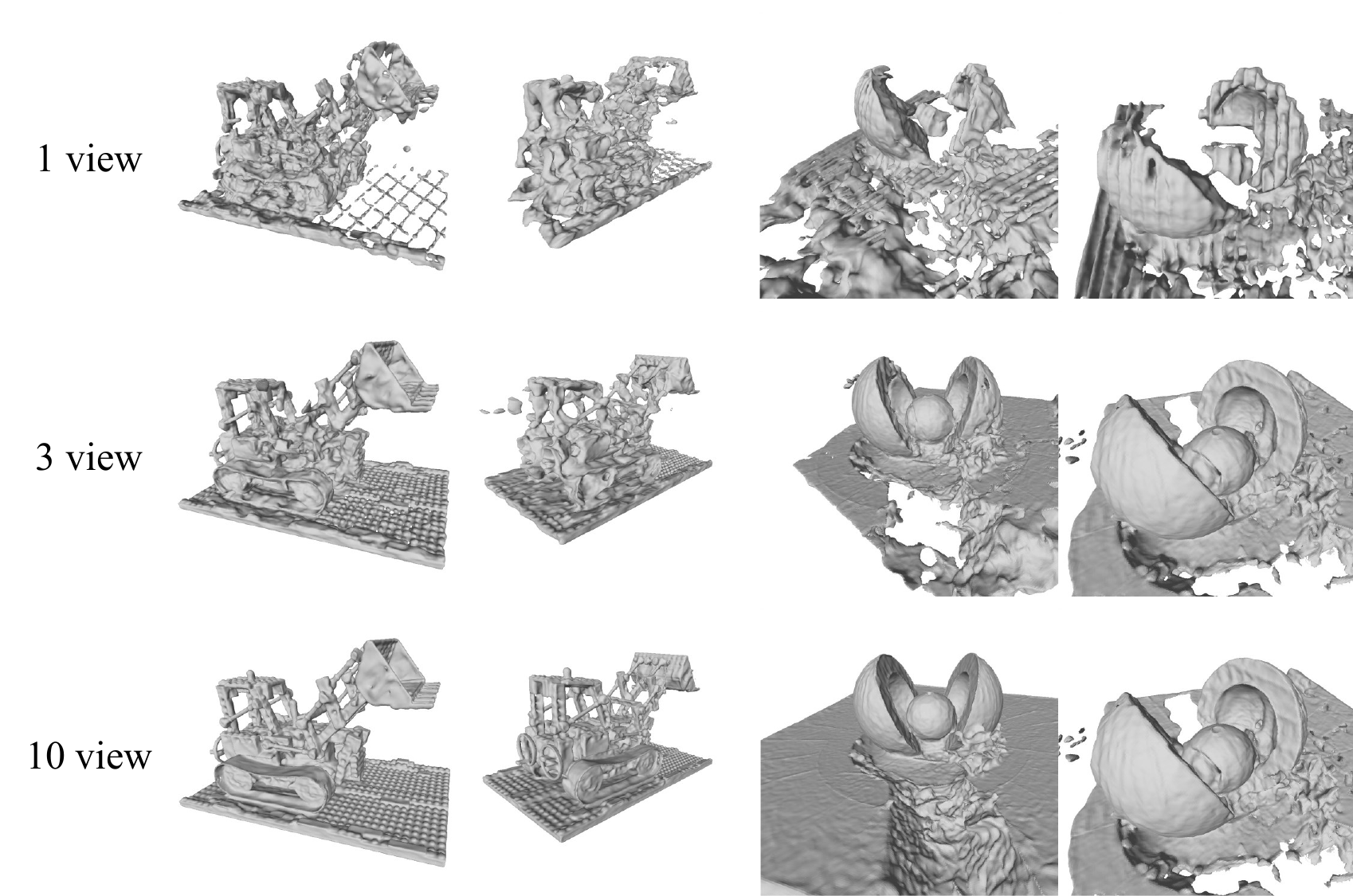}
        \subcaption{Grid-pattern (single pattern per view)}
 \vspace{0.1cm}
    \end{minipage}
    \begin{minipage}[b]{1\linewidth}
        \centering
        \includegraphics[width=\columnwidth]{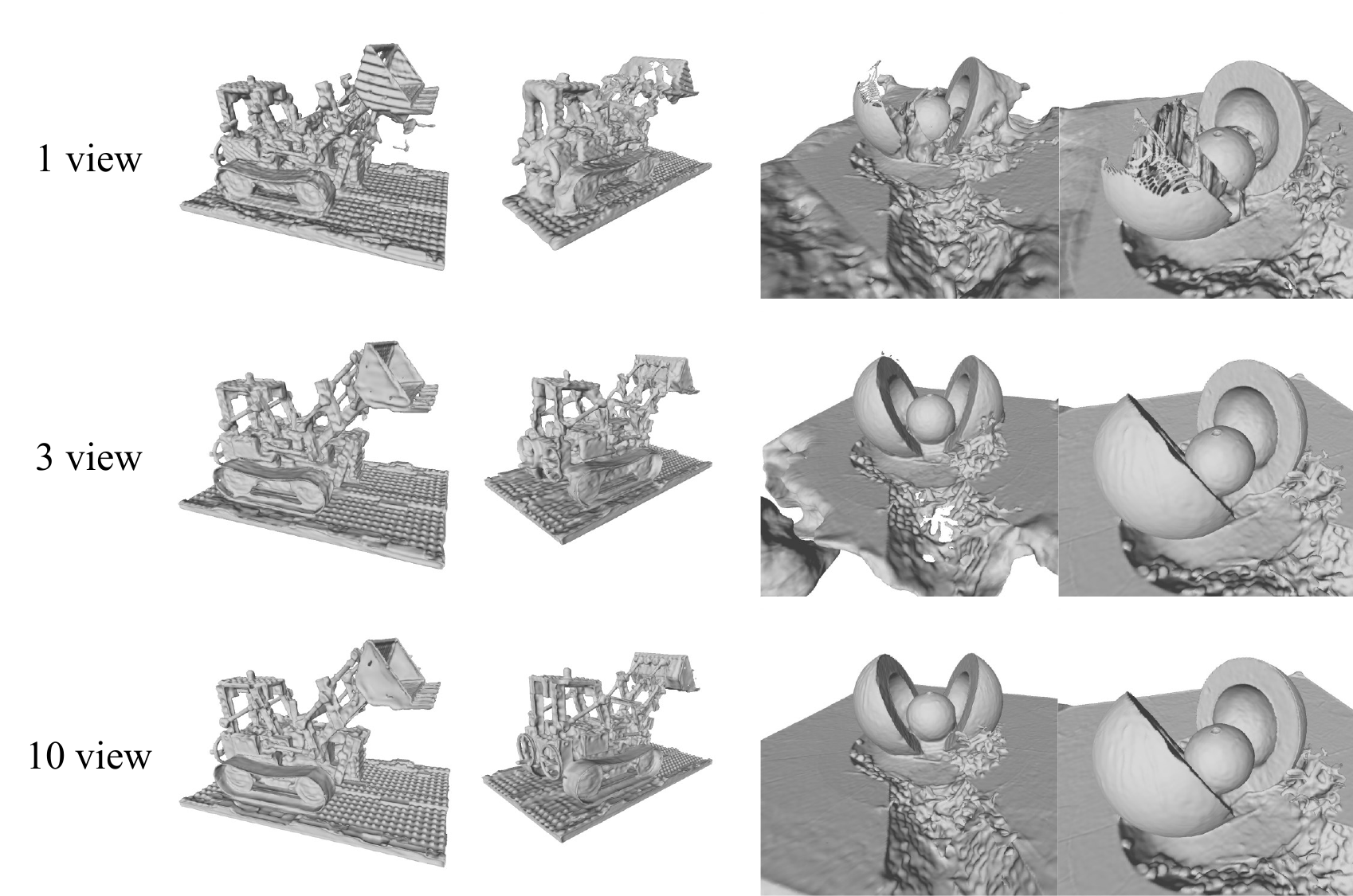}
        \subcaption{Colorful line pattern (single pattern per view)}
  \vspace{-0.2cm}
    \end{minipage}
    \caption{The reconstructed shapes with various patterns and number of viewpoints. \textbf{Left}: NeRF-Synthetic (Lego). \textbf{Right}: BlendedMVS (Stone).}
    \label{fig:various_patterns}
\end{figure}

We further investigated the effects of changing projected patterns from the Cross-laser pattern.
In this experiment, we prepared synthetic datasets with new patterns such as Graycode (20 patterns), Grid-pattern (One-shot), and Colorful line pattern (One-shot) as shown in \autoref{fig:dataset_various_patterns}.
While Graycode requires multiple shots for one viewpoint, Grid and Colorful line only require one shot for each viewpoint, in exchange for an increase in the difficulty of finding correspondence.
All these patterns were projected by a single virtual projector with a baseline length $60cm$.
We trained the networks with a different number of viewpoints (1, 3, 10) to examine if it is possible to acquire 3D shapes with a limited number of viewpoints (As for Graycode, we could not train the networks in 3 viewpoints and 10 viewpoints due to VRAM limitation).
Although we also trained the networks without pattern projection in 3 viewpoints by NeuS, we could not get any meaningful results.

\autoref{fig:various_patterns} shows the reconstructed shapes with the aforementioned patterns and the number of viewpoints.
It seems to be difficult to reconstruct a complex shape with fewer viewpoints, still, we can see approximate shapes even on 1 viewpoint cases except Grid-pattern.
Some results on NeRF-Synthetic have quantization error, but this is because data synthesis caused quantization of pattern projection due to their Ground-truth depth data, as it is provided in 8-bits image format.
On the other hand, the results on BlendedMVS do not have such errors as their Ground-truth depth data is provided in PFM format.
Overall, we confirmed that One-shot / single-viewpoint (only 1 training image) 3D shape reconstruction is possible with the proposed method.

\subsection{Demonstration with real data}

\begin{figure}[t!]
    \begin{minipage}[b]{0.32\linewidth}
        \centering
        \includegraphics[width=\columnwidth]{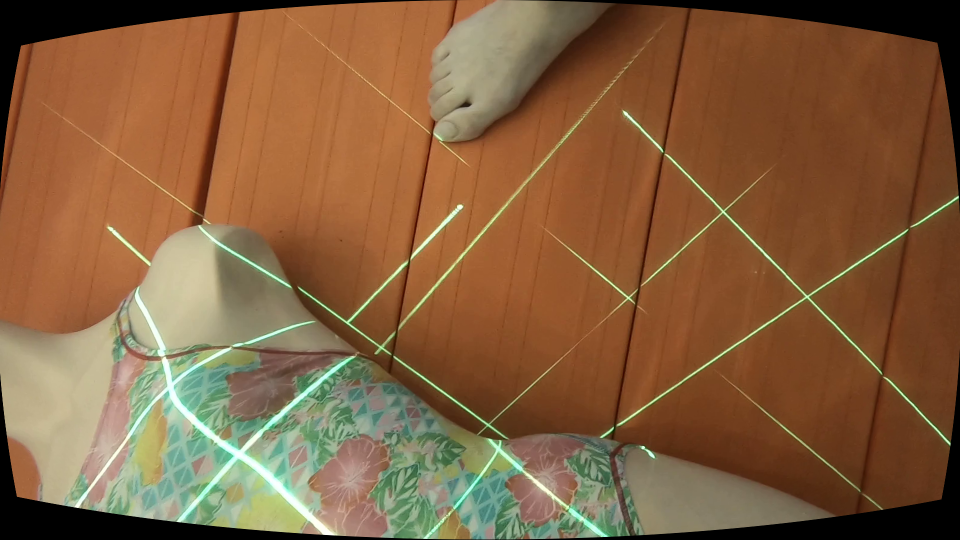}
        \end{minipage}
    \begin{minipage}[b]{0.32\linewidth}
        \centering
        \includegraphics[width=\columnwidth]{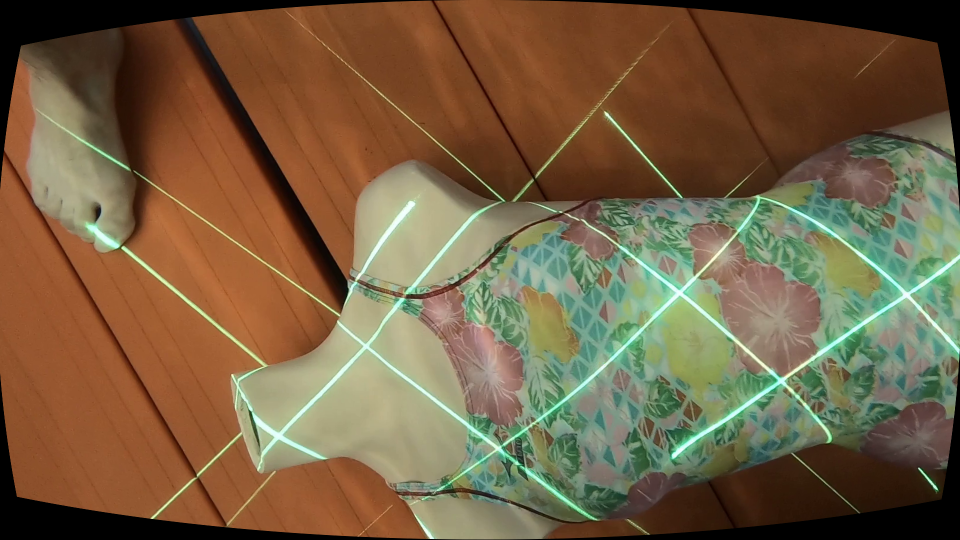}
    \end{minipage}
    \begin{minipage}[b]{0.32\linewidth}
        \centering
        \includegraphics[width=\columnwidth]{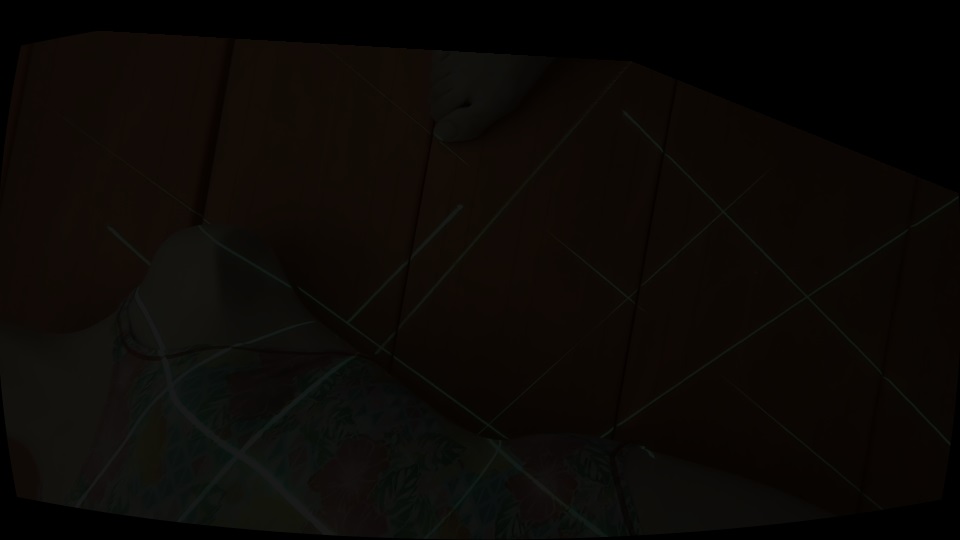}
    \end{minipage}
    \vspace{-0.3cm}
    \caption{Example images of the real data captured underwater, where contrast was decreased by scattering effect. The rightmost one is an image of Dark illumination condition; note that it looks almost black for human eyes.}
    \label{fig:mannequin_image}
\end{figure}

\begin{figure*}[t!]
    \begin{minipage}[b]{0.19\linewidth}
        \centering
        \includegraphics[width=\columnwidth]{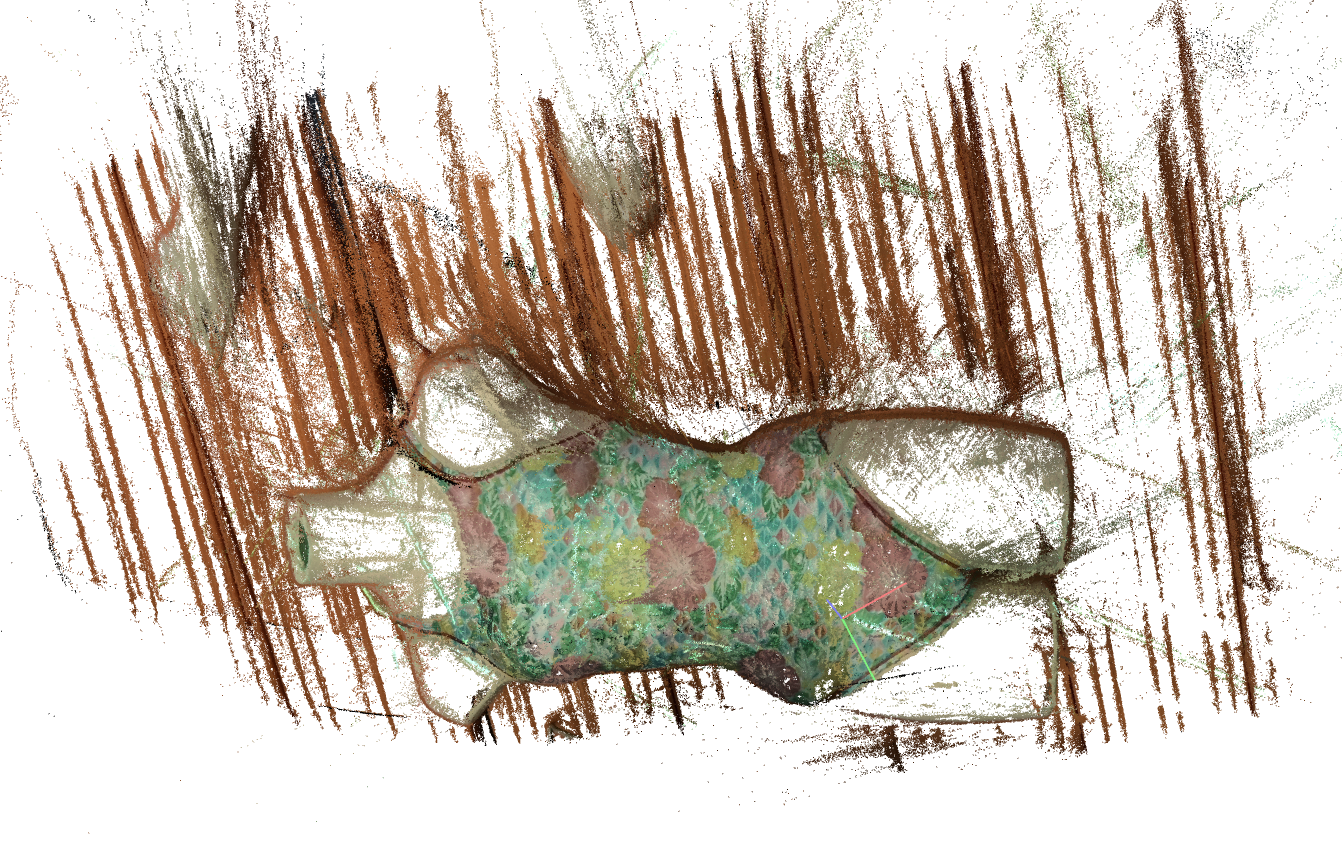}
    \end{minipage}
    \begin{minipage}[b]{0.19\linewidth}
        \centering
        \includegraphics[width=\columnwidth]{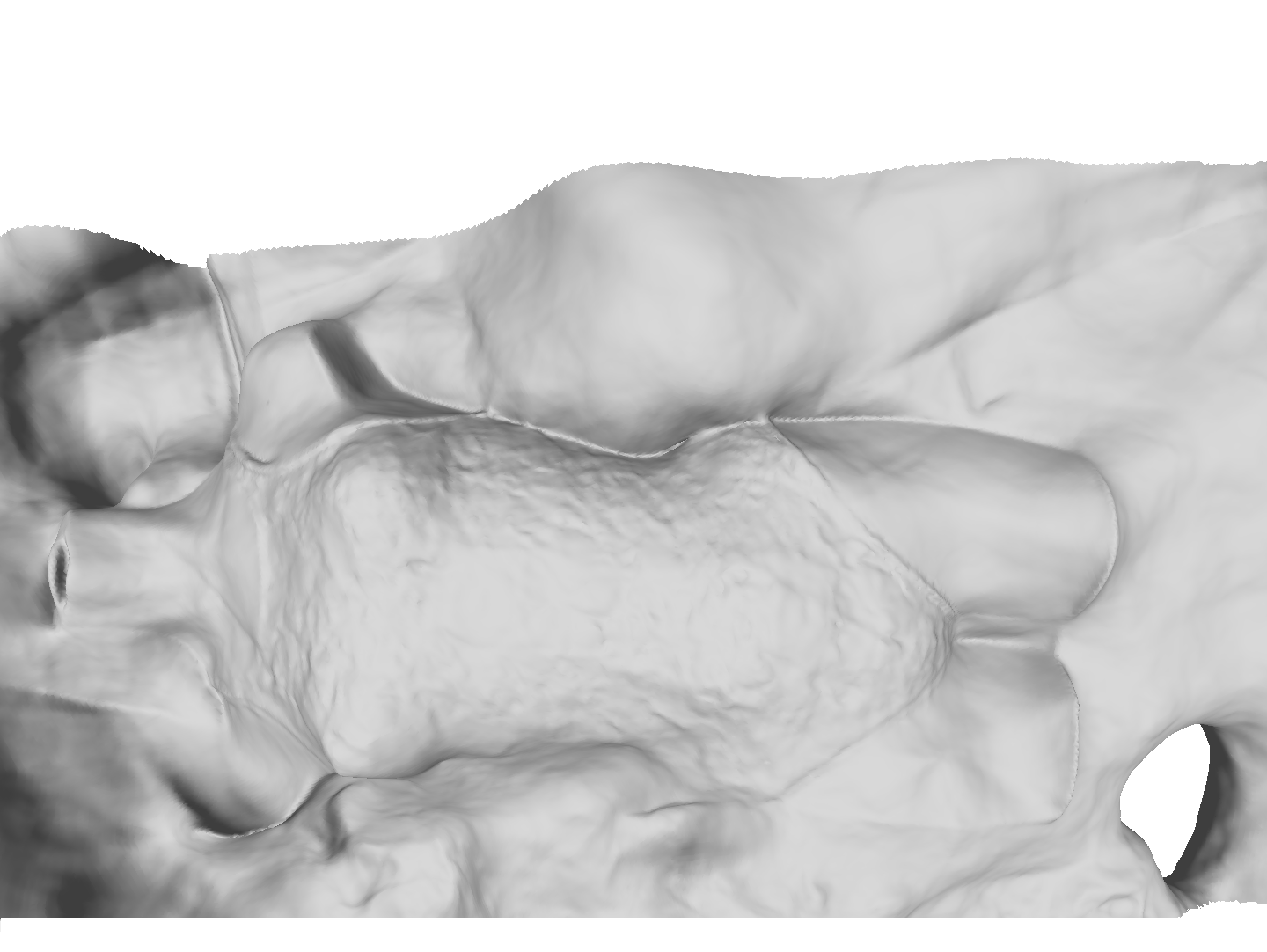}
    \end{minipage}
    \begin{minipage}[b]{0.19\linewidth}
        \centering
        \includegraphics[width=\columnwidth]{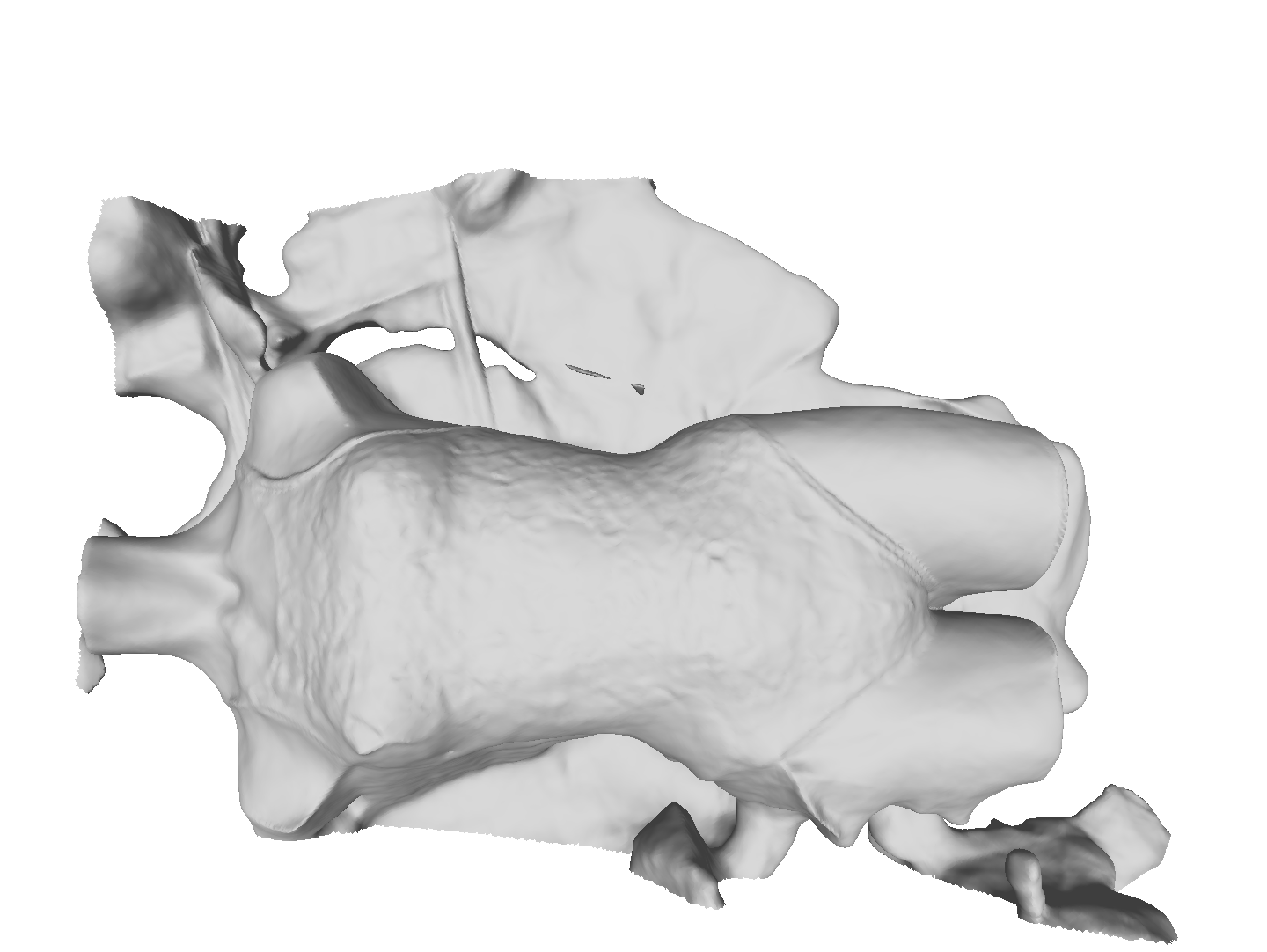}
    \end{minipage}
    \begin{minipage}[b]{0.19\linewidth}
        \centering
        \includegraphics[width=\columnwidth]{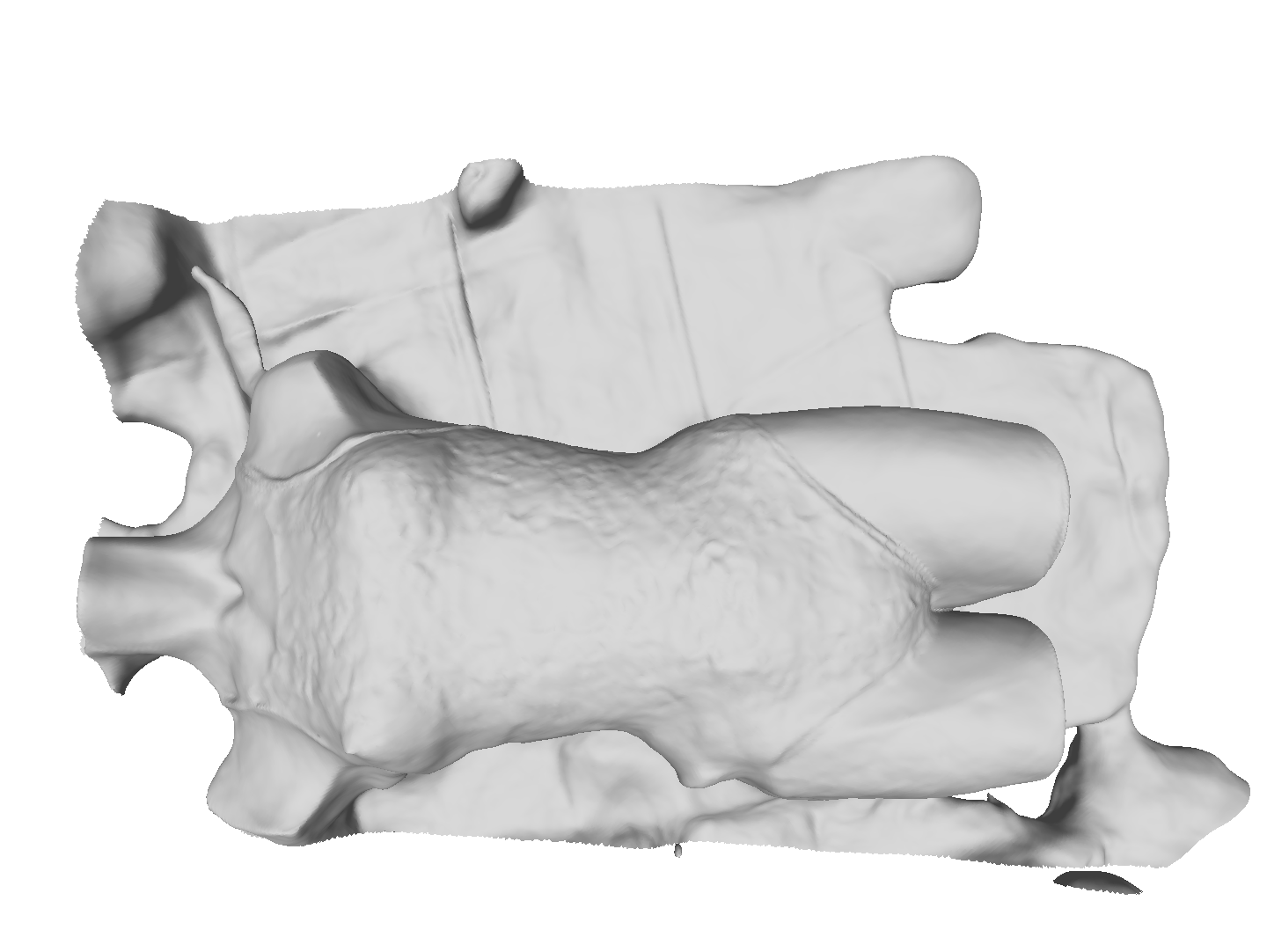}
    \end{minipage}
    \begin{minipage}[b]{0.19\linewidth}
        \centering
        \includegraphics[width=\columnwidth]{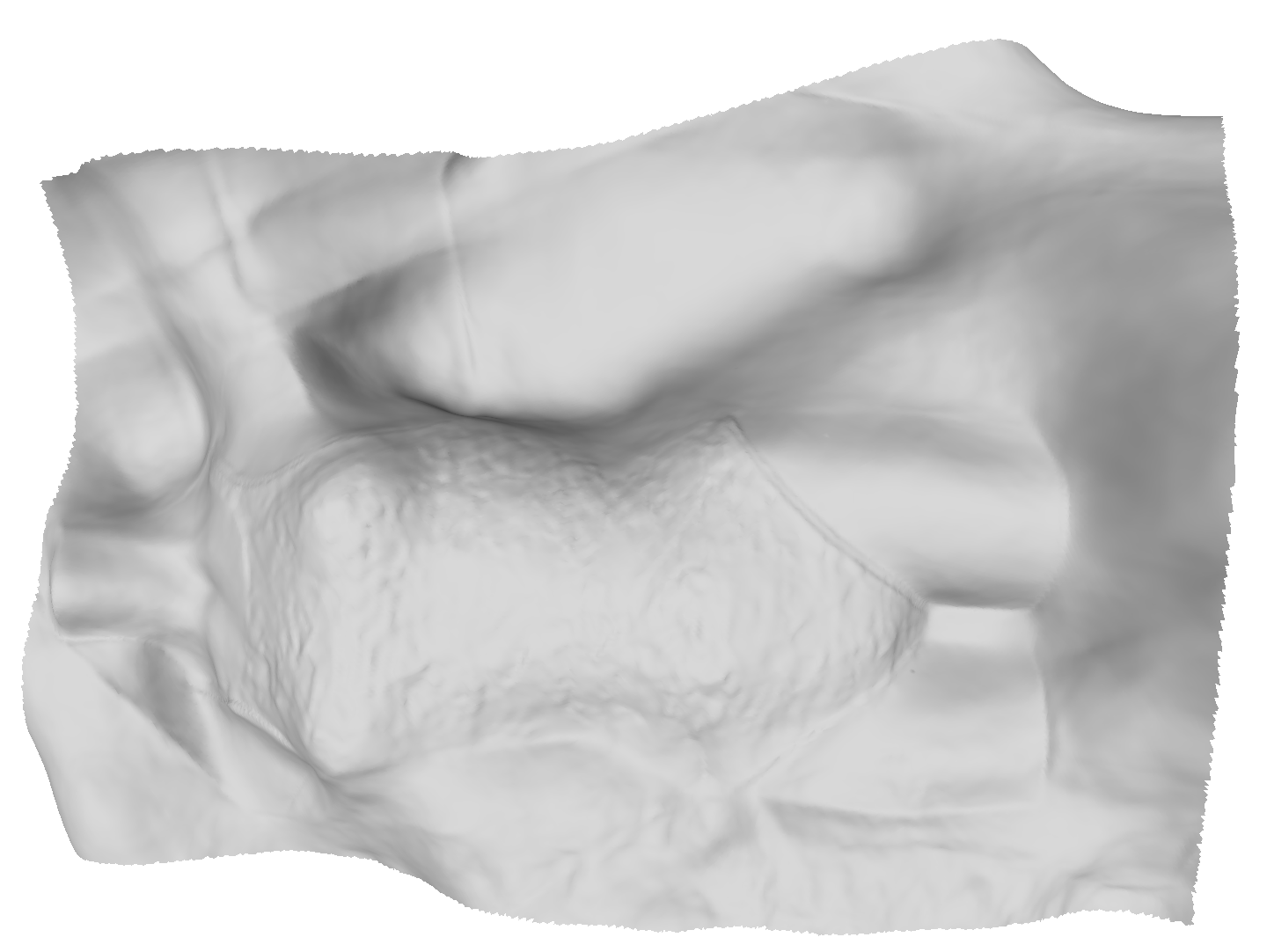}
    \end{minipage}
    \begin{minipage}[b]{0.19\linewidth}
        \centering
        \includegraphics[width=\columnwidth]{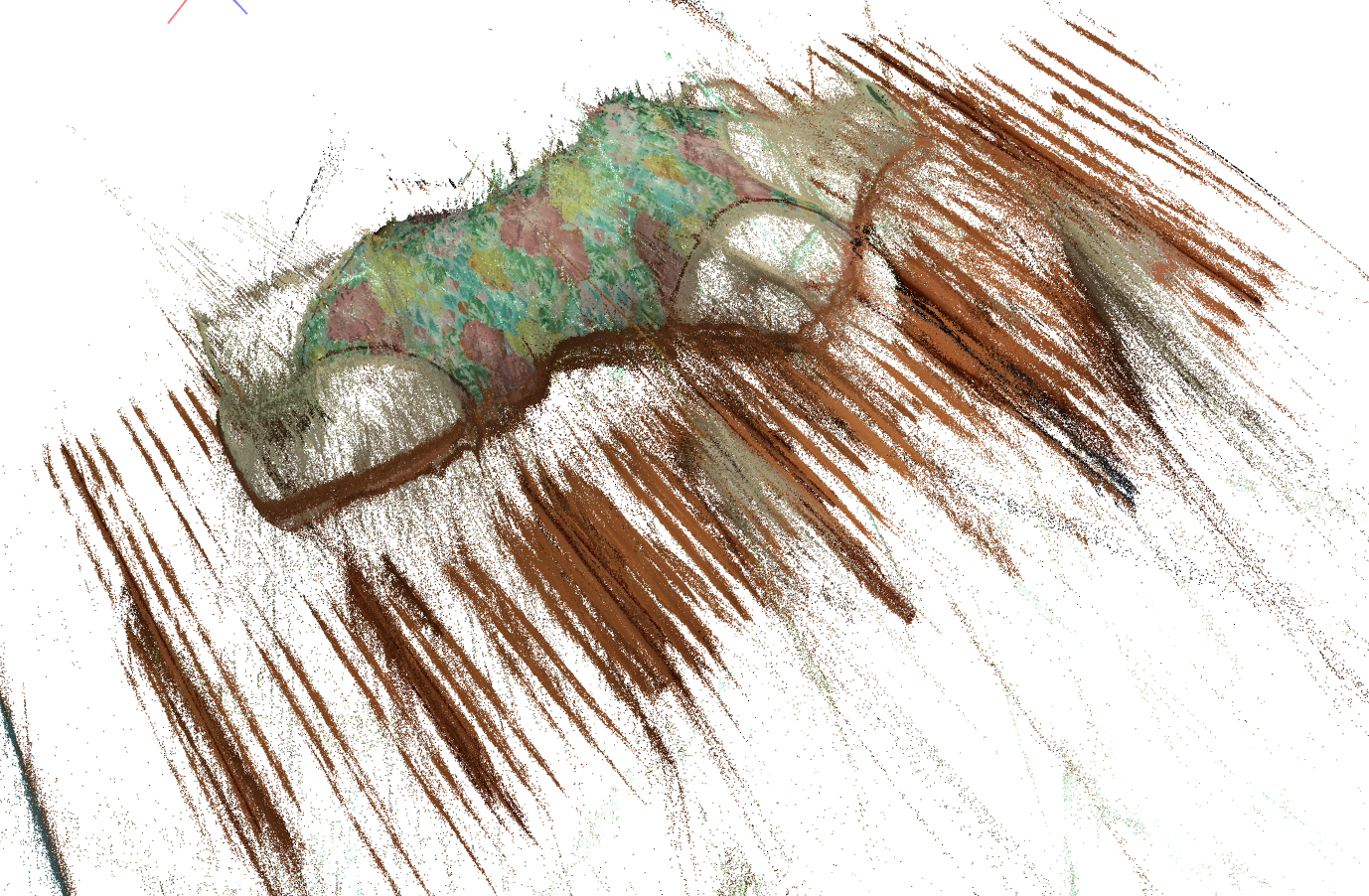}
        \subcaption{COLMAP}
    \end{minipage}
    \begin{minipage}[b]{0.19\linewidth}
        \centering
        \includegraphics[width=\columnwidth]{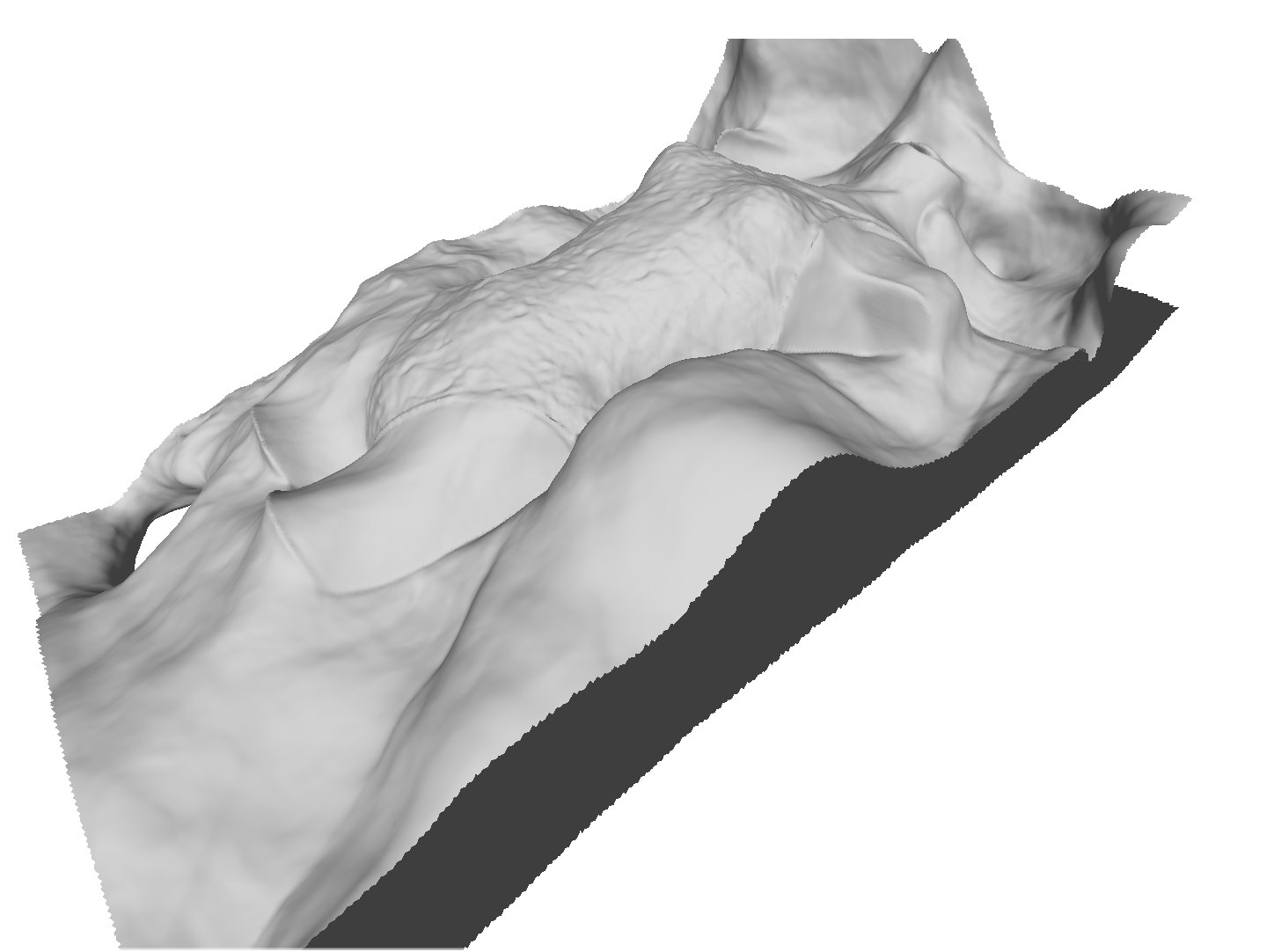}
        \subcaption{NeuS}
    \end{minipage}
    \begin{minipage}[b]{0.19\linewidth}
        \centering
        \includegraphics[width=\columnwidth]{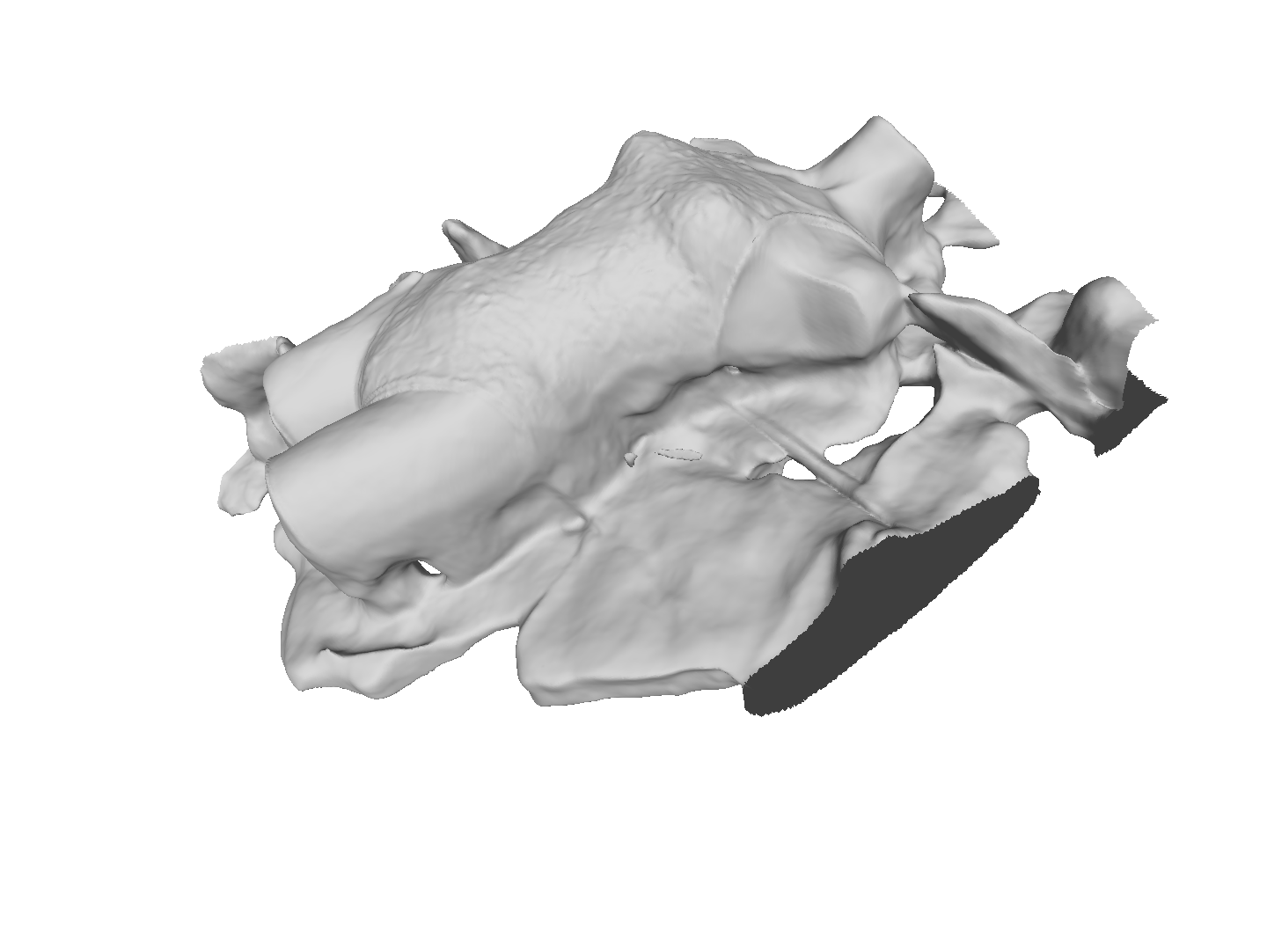}
        \subcaption{w/o Illum. estimation}
    \end{minipage}
    \begin{minipage}[b]{0.19\linewidth}
        \centering
        \includegraphics[width=\columnwidth]{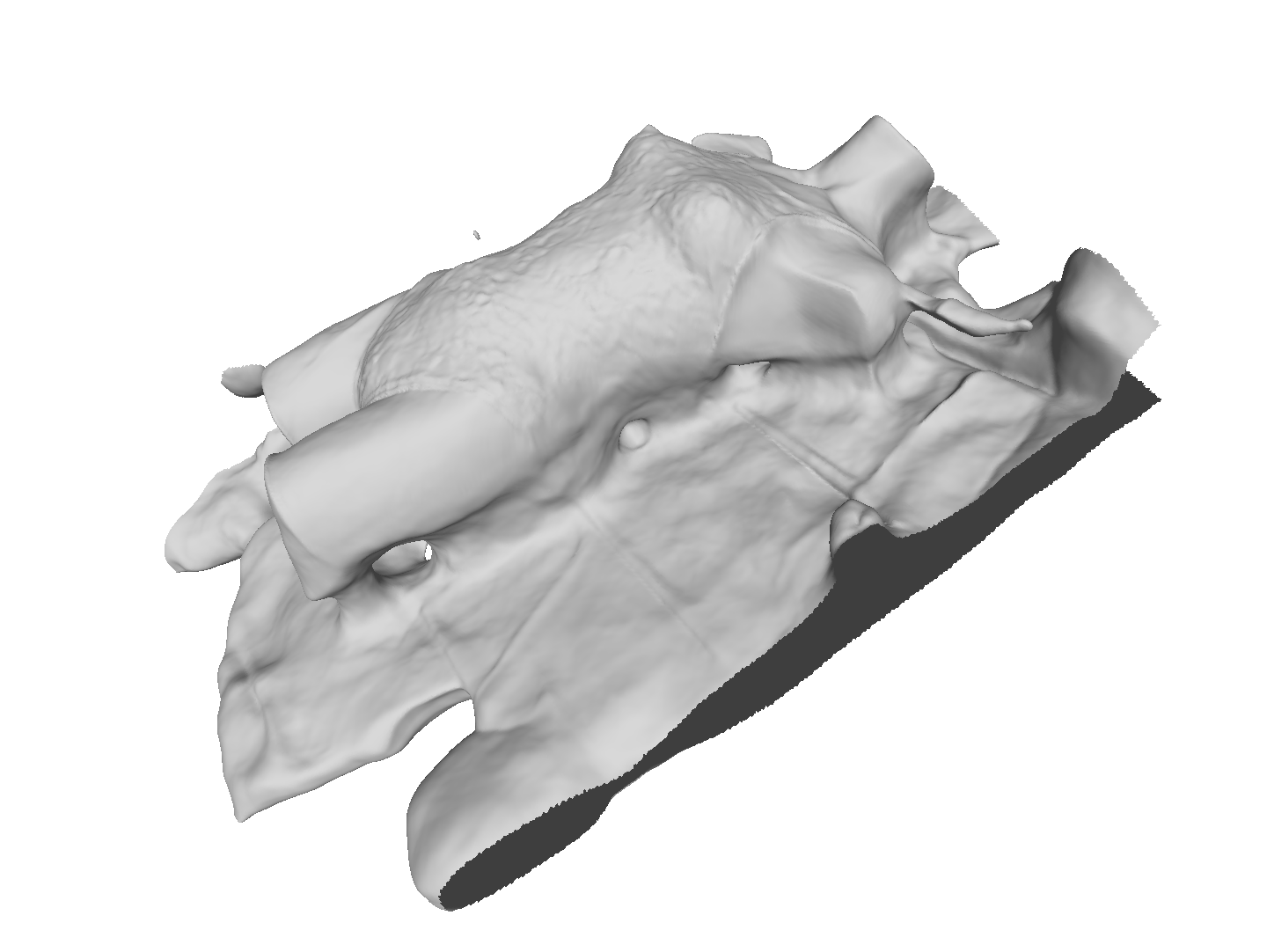}
        \subcaption{w Illum. estimation}
    \end{minipage}
    \begin{minipage}[b]{0.19\linewidth}
        \centering
        \includegraphics[width=\columnwidth]{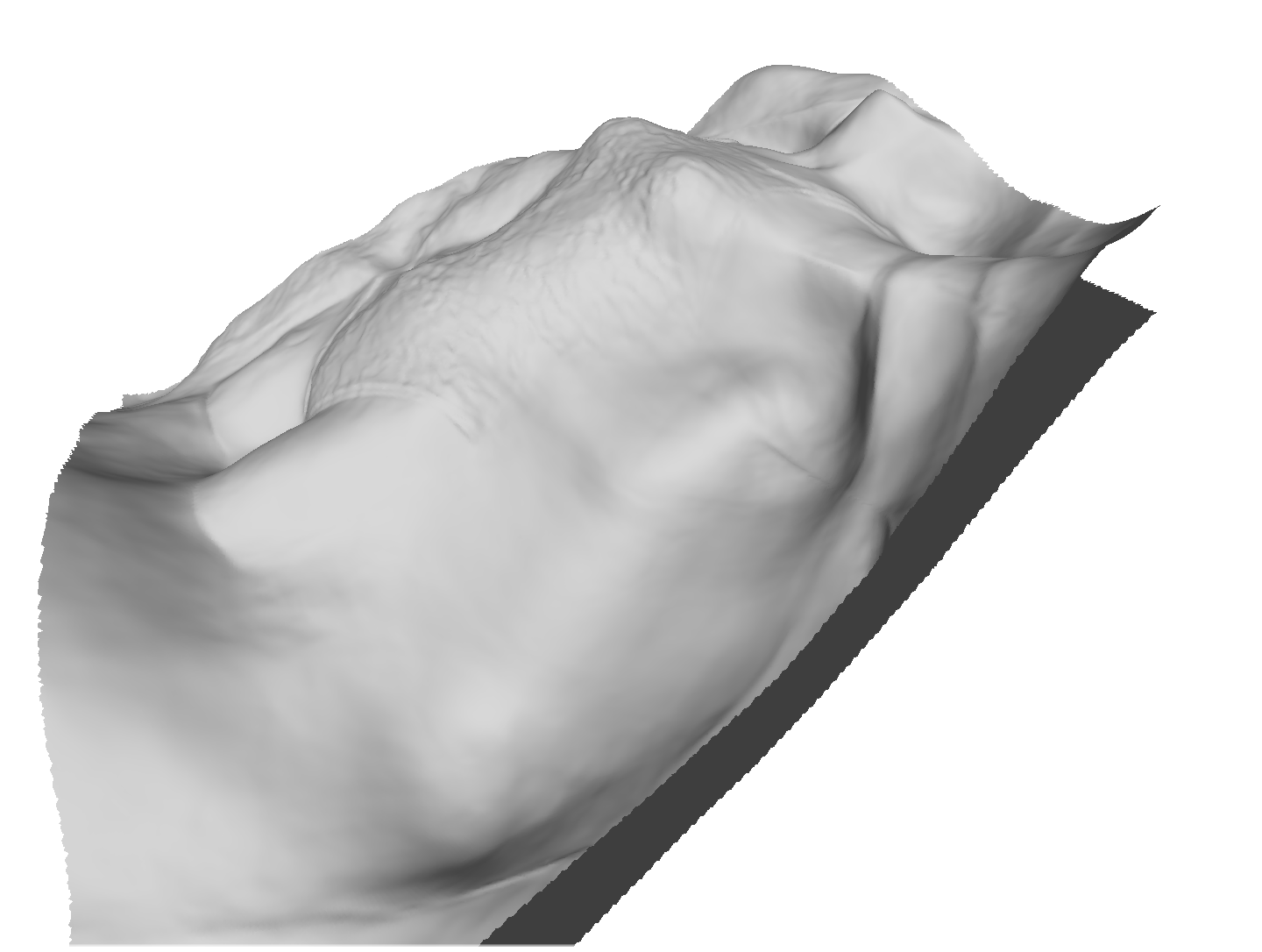}
        \subcaption{Dark illum.}
    \end{minipage}
    \caption{The reconstructed shapes of the real data.}
    \label{fig:mannequin_reconst}
\end{figure*}

\begin{figure}[t!]
    \centering
    \begin{minipage}[b]{0.42\linewidth}
        \centering
        \includegraphics[width=\columnwidth]{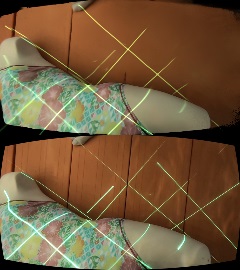}
        \subcaption{w/o Illumination estimation}
        \end{minipage}
    \begin{minipage}[b]{0.42\linewidth}
        \centering
        \includegraphics[width=\columnwidth]{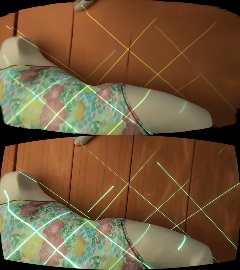}
        \subcaption{w Illumination estimation}
    \end{minipage}
    \caption{Comparison of the rendered images with and without illumination estimation. The top row is the rendered images and the bottom row is the Ground-truth images. Patterns around top-right region disappeared ``without illumination estimation'' scenario.}
    \label{fig:mannequin_illum_comparison}
\end{figure}

\begin{figure}[t!]
    \centering
    \begin{minipage}[b]{0.42\linewidth}
        \centering
        \includegraphics[width=\columnwidth]{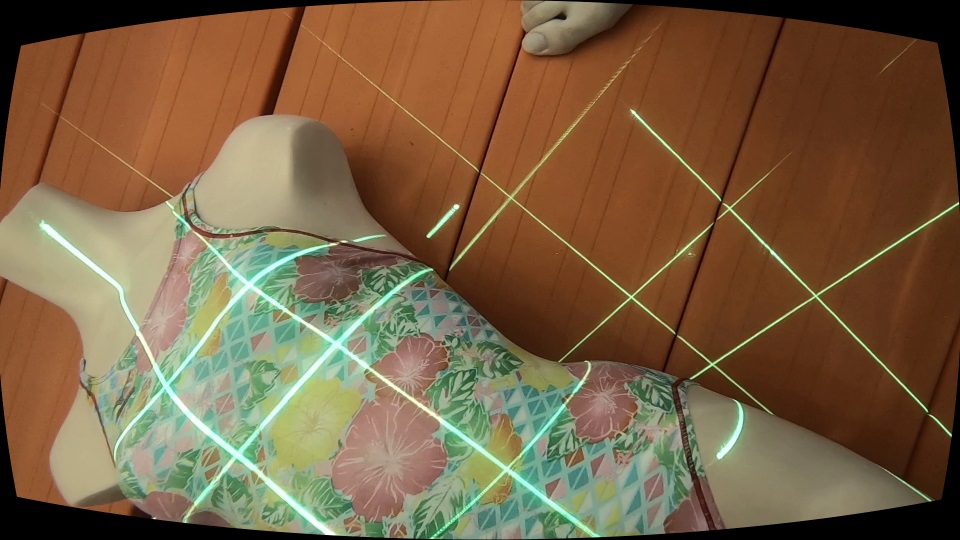}
        \subcaption{Original image}
        \end{minipage}
    \begin{minipage}[b]{0.42\linewidth}
        \centering
        \includegraphics[width=\columnwidth]{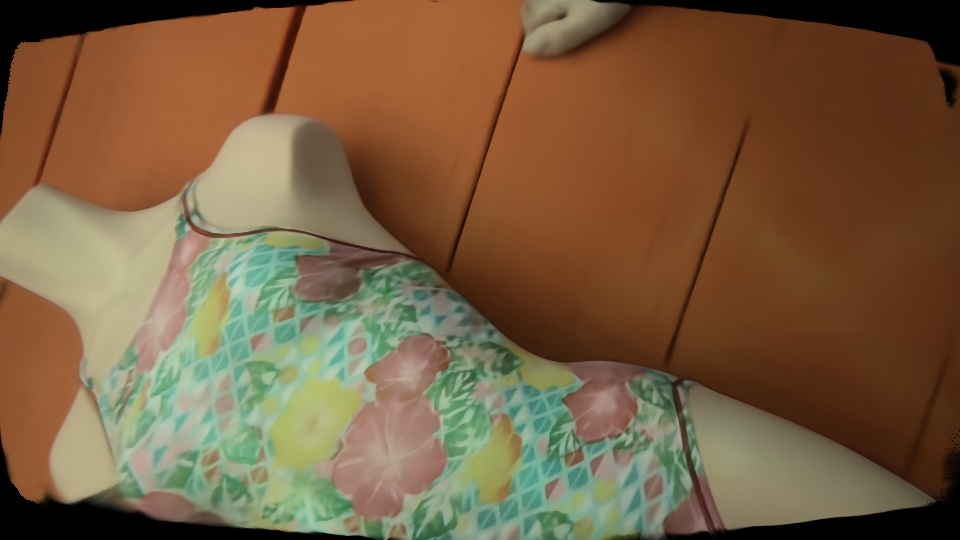}
        \subcaption{Texture retrieval image}
    \end{minipage}
    \caption{Texture retrieval results of the real data.}
    \label{fig:mannequin_texture}
\end{figure}

In order to confirm the feasibility of the proposed method in real environments, we conducted experiments with real data captured underwater, where scenes are usually low-texture/low-contrast with strong scattering effects.
Since structured light techniques have been conducted for more than several decades, we follow the common technique to achieve accurate projector calibration~\cite{ActiveLight:Ikeuchi20:Springer}.
We used a Remotely Operated Vehicle (ROV) in the pool 
to capture an underwater scene with a table and a mannequin wearing swimwear.
The ROV is equipped with 4 green cross-laser projectors, which were calibrated beforehand.
\autoref{fig:mannequin_image} shows example images of the real data.

Since the captured images contain distortion due to refraction on the waterproof housing, we removed it by approximating refraction to radial lens distortion.
Afterward, we used COLMAP to obtain the system transformations in time series, as well as the 3D shape reconstructed by conventional algorithms\cite{schoenberger2016mvs,schoenberger2016sfm}.
As for the projector calibration, we used a technique proposed in \cite{Wang2022ICIP} to get the laser plane parameters.
Then, we manually set the field-of-view according to the catalog specification and adjusted their positions along the depth-axis.
Note that, the refraction of the projectors can be ignored except change of field-of-view because the laser lines always lay on the refractive plane.

\autoref{fig:mannequin_reconst} shows the reconstruction results of the real data.
The proposed method achieved dense and accurate reconstruction compared to COLMAP and NeuS, especially in texture-less regions where active stereo is advantageous.
As for quantitative error, chamfer distance between the ground-truth (the mannequin shape captured by KinectFusion~\cite{KinectFusion}) was $3.645mm$ for NeuS, while \methodName achieved $2.258mm$.
Note that the projected pattern may be disturbance for NeuS, but we empirically observed that projected pattern can be consumed as specularity by the Color network.
Since we do not have Ground-truth illumination parameters for this data, the reconstructed shape is a bit collapsed without illumination estimation scenario.
By estimating illumination, the reconstructed shape is refined, showing the effectiveness of the proposed method.

To analyze the effect of illumination estimation, we compared rendered images with and without illumination estimation in \autoref{fig:mannequin_illum_comparison}.
As can been seen, some laser lines disappeared due to wrong illumination parameters in ``without illumination estimation'' case, while more laser lines are properly projected in ``with illumination estimation'' case.

In addition, we also conducted an evaluation in the Dark illumination condition.
Since it is difficult to create a real dark illumination environment (as the ROV may cause an accident), we simply decreased the brightness  as shown in \autoref{fig:mannequin_image} (Right).
The reconstructed shape is shown in \autoref{fig:mannequin_reconst} (d), where approximate reconstruction is still possible, even though the details of the shapes are missing.

Finally, \autoref{fig:mannequin_texture} shows texture recovery results by using properly estimated illumination parameters.
It is proved that the proposed method is capable of accurate texture retrieval with almost no artifacts on the real data.

%% file: sec/6_conclusion.tex
\section{Conclusion}
\label{sec:conclusion}

In this paper, we propose Neural SDF for active stereo systems to enable implicit correspondence search and triangulation in generalized SL.
With our method, texture-less/low-texture regions by low-light condition are compensated by SL, thereby, shapes are successfully reconstructed even with a small number of inputs.
By comprehensive experiments, it is proved that the proposed method is advantageous compared to the ordinary Neural SDF under 
low illumination or a small number of input images.
It is also demonstrated that the proposed method successfully worked in a real underwater scenario.
%
As future work, simultaneous optimization of system pose and shape as well as 
handling inter-reflections of the projected laser pattern are interesting.